\pdfoutput=1
\documentclass[11pt]{article}

\usepackage[T1]{fontenc}
\usepackage[utf8]{inputenc}
\usepackage{cite}
\usepackage{amsmath,amssymb,amsfonts}
\usepackage{algorithm}
\usepackage{algorithmic}
\usepackage{graphicx}
\usepackage{multirow}
\usepackage{textcomp}
\usepackage{booktabs}
\usepackage[table]{xcolor}
\usepackage{subcaption}
\usepackage{array}
\usepackage{xspace}
\usepackage{hyperref}
\hypersetup{
  colorlinks=true,
  linkcolor=blue,
  citecolor=blue,
  urlcolor=blue
}

\newcommand{\ie}{\textit{i.e.,}\xspace}

\newcommand{\methodname}{\textsc{VIPP-SR}\xspace}
\newcommand{\vipgan}{\textsc{VIP-GAN}\xspace}
\newcolumntype{Y}{>{\centering\arraybackslash}p{0.098\textwidth}}
\newcolumntype{M}{>{\raggedright\arraybackslash}p{0.135\textwidth}}
\newenvironment{keywords}{%
  \par\smallskip\noindent\textbf{Keywords: }\ignorespaces
}{\par\smallskip}

\def\BibTeX{{\rm B\kern-.05em{\sc i\kern-.025em b}\kern-.08em
    T\kern-.1667em\lower.7ex\hbox{E}\kern-.125emX}}
\begin{document}

\title{Anatomically Consistent Cross-Contrast Super-Resolution of
 Anisotropic Brain T2w MRI}

\author{%
Mengqi Shen$^{1}$, Haicheng Wang$^{4}$, Meghna Trivedi$^{3}$,
Tony J.~Wang$^{2}$,\\
Yuanguang Xu$^{2}$, Yingyan Zeng$^{4}$, and Yading Yuan$^{1,2}$\\[0.6em]
\small $^{1}$Data Science Institute, Columbia University, New York, NY, USA\\
\small $^{2}$Department of Radiation Oncology, Columbia University Irving Medical Center,\\
\small New York, NY, USA\\
\small $^{3}$Herbert Irving Comprehensive Cancer Center, Columbia University,\\
\small New York, NY, USA\\
\small $^{4}$Department of Mechanical and Materials Engineering, University of Cincinnati,\\
\small Cincinnati, OH, USA\\[0.4em]
\small\texttt{ms7001@columbia.edu}; \texttt{yading.yuan@columbia.edu}\\
\small\texttt{tjw2117@cumc.columbia.edu}; \texttt{yx2010@cumc.columbia.edu}\\
\small\texttt{mst2134@cumc.columbia.edu}; \texttt{wang4hc@mail.uc.edu}\\
\small\texttt{zengyy@ucmail.uc.edu}
}
\date{}

\maketitle
\pagenumbering{arabic}

% =====================================================================
\begin{abstract}
T2-weighted (T2w) brain MRI provides fluid-sensitive soft-tissue
contrast that is important for neuro-oncology and radiotherapy
planning. However, T2w scans are acquired with anisotropic
voxels and appear blurred or stair-stepped on coronal and sagittal views, which obscures small structures and weakens any downstream
3D analysis. We propose \methodname (View-Independent Patched Projection Super-Resolution), a cross-contrast guided
super-resolution framework that restores the inter-plane resolution
of an existing anisotropic T2w volume without an
isotropic ground-truth T2w. 

\methodname first trains a view-independent patched generator
(\vipgan) to learn local T1c-to-T2w anatomical correspondence from
high-resolution axial slices. The trained generator is then applied
to axial, coronal, and sagittal views of the T1c volume to
generate three orthogonal T2w estimates. Shape-preserving patching and
deepest-skip removal reduce view-specific shortcuts, thereby constraining the generator to learn patch-local representations and enabling the
zero-shot inter-plane transfer.
Central to \methodname, a projection-based optimization then enforces
anatomical consistency across the three view-specific volumes, fusing
them by balancing inter-plane self-consistency against per-view data
fidelity.
The generator is trained on BraTS-MET and evaluated on both the held-out BraTS-MET testing set and 
the BraTS-GLI cohort without retraining, assessing the cross-cohort generalizability. 
The results validate that \methodname improves downstream segmentation over the real anisotropic T2w baseline---raising mean-label Dice from $0.330$ to $0.465$ on BraTS-MET and, zero-shot, from $0.473$ to $0.563$ on BraTS-GLI---and ablation studies identify inter-plane self-consistency as the main source of the gain.
\end{abstract}


% =====================================================================
\section{Introduction}
\label{sec:intro}
% =====================================================================

T2-weighted (T2w) brain Magnetic resonance imaging (MRI) provides
fluid-sensitive soft-tissue contrast that is important in
neuro-oncology and radiotherapy planning, particularly for
depicting peritumoral edema, gliosis, and postoperative resection cavities.
In routine clinical protocols\cite{teyateeti2020brain, castellano2021advanced}, 
however, T2w images are often
acquired as two-dimensional fast spin-echo stacks with
anisotropic voxel spacing. The intra-plane resolution on the
acquisition plane, commonly axial, is typically near
millimetre or sub-millimetre, whereas the inter-plane
slice thickness or spacing is often several millimetres
(e.g., 3-5\,mm). This reflects the practical trade-off among
scan time, signal-to-noise ratio, motion sensitivity, and
patient throughput~\cite{deoni2022simultaneous,
sui2021gradient,uus2025scanner}. 
Such volumes can appear
sharp on the native acquisition plane but blurred or
stair-stepped on coronal and sagittal views. The resulting
loss of inter-plane high-frequency (\ie detailed)
information can obscure small structures, bias volumetric
measurements, and degrade downstream 3D analysis built
for isotropic anatomy~\cite{liu20213d,sui2021gradient}.

% \textbf{Super-resolution rather than synthesis from scratch.}
A natural temptation is to synthesize
an isotropic T2w volume directly from the co-acquired isotropic T1c.
However, the mapping from T1c to T2w is not deterministic as 
the two modalities encode different relaxation and enhancement mechanisms, 
and a T2w generated from T1c alone risks
pathology-specific hallucinations~\cite{cohen2018distribution,iglesias2023synthsr}. 
We therefore frame the task as super-resolution (SR) rather than
synthesis. The patient's own anisotropic T2w is already available, and
our goal is to recover its inter-plane detailed anatomical information while staying anchored
to the patient-specific anatomy that T1c reveals at isotropic
resolution. This becomes a cross-contrast guided super-resolution task,
where the reference T1c provides anatomical structure and the acquired
T2w provides the contrast prior.

Existing work on anisotropic MRI reconstruction can be grouped into three
directions. 
Classical model-based reconstruction formulates SR as an inverse problem with an explicit forward model for slice acquisition, image
degradation, motion, and downsampling~\cite{jiang2007svr,gholipour2010robust}.
Its reliability depends on the accuracy of the assumed acquisition model, which can vary
across scanners and protocols.
Supervised deep super-resolution learns
a low-resolution (LR) to high-resolution (HR) mapping from paired training
data~\cite{chen2018dcsrn,lyu2020multi,liu20213d,kim2024adaptive,
lin2023sptsr}.
These methods can be effective when matched high-resolution data are available, but
isotropic high-resolution T2w is rarely available in retrospective
brain tumor cohorts.
Self-supervised and generative-prior SR remove the paired
requirement using within-volume redundancy~\cite{zhao2020smore},
cross-orientation consistency~\cite{benisty2024simple} or latent
diffusion priors~\cite{wang2023inversesr,pinaya2022ldmbrain}.
However, they
operate slice by slice and enforce 3D consistency only implicitly, so
independent per-view estimates of the same anatomy can disagree
off-plane, leaving the reconstructed volume anatomically inconsistent
across views.

\begin{figure}[t]
  \centering
  \includegraphics[width=0.84\textwidth]{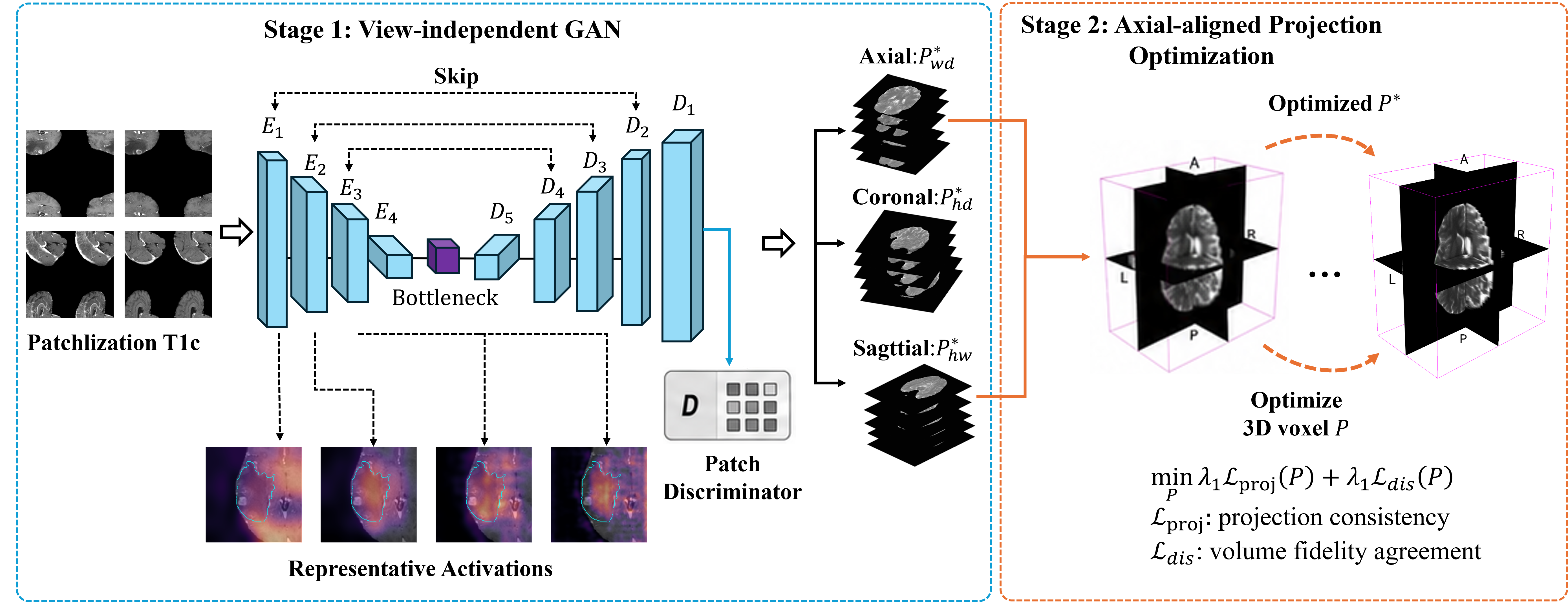}
  \caption{Overview of the proposed \methodname pipeline. Stage~1 trains
  a view-independent patched GAN on axial T1c$\rightarrow$T2w pairs and
  applies the frozen generator along the three orthogonal axes. Stage~2
  fuses the resulting view-specific volumes with an axial-aligned
  projection objective to produce the final isotropic T2w estimate.}
  \label{fig:overview}
\end{figure}

To address the aforementioned challenges, we exploit the observation that the axial T2w
slices retain high intra-plane anatomical detail and are spatially paired
with the corresponding T1c slices.
 The key idea is to use T1c as a high-resolution anatomical reference
without treating it as a replacement for the acquired T2w. In this way,
the method can exploit the isotropic structure in T1c while remaining
anchored to the patient's own anisotropic T2w contrast.
The proposed \methodname (View-Independent Patched Projection Super-Resolution) operates in two stages as shown in Fig.~\ref{fig:overview}.
In Stage~1, a view-independent patched generator, \vipgan (View-Independent Patched Generative Adverisal Network), is proposed and trained
on high-quality axial T1c$\to$T2w pairs. The generator is designed to
learn local cross-contrast anatomical correspondence rather than
memorize a single viewing direction. It is therefore used on axial, coronal, and sagittal views of the same T1c
volume, producing three orthogonal T2w estimates. Because each view is
generated independently, these estimates capture complementary
intra-plane detail but need not agree off their own plane, so
reconciling them into a single anatomically consistent 3D volume is
essential.
In Stage~2, these estimated volumes are fused through projection-based
optimization. The optimization treats the view-specific outputs as
complementary constraints on one unknown 3D volume. It balances
self-consistency, which enforces anatomical consistency across the
three view-specific volumes, with data fidelity, which keeps the result
close to the generated T2w evidence. The full framework does not require isotropic ground-truth
T2w at any stage.

\textbf{Contributions.} First, we propose a two-stage T1c-guided T2w
super-resolution framework that uses T1c as a high-resolution
anatomical reference while remaining anchored to the acquired
anisotropic T2w contrast. Second, we design a view-independent patched
generator, \vipgan, whose combination of shape-preserving patching, brain-mask
conditioning, and deepest-skip removal lets a generator trained on
axial slices transfer zero-shot to the coronal and sagittal planes. Third, and central to this work, we introduce a projection-based 3D
optimization that enforces anatomical consistency across the three
view-specific volumes, fusing them by balancing inter-plane
self-consistency against per-view data fidelity. Fourth, we validate
the framework through controlled ablations and cross-cohort evaluation,
showing that this inter-plane anatomical consistency is the primary
source of the downstream gain and that a generator in Stage 1 trained
on BraTS-MET transfers to BraTS-GLI without retraining.

\section{Related Work}
\label{sec_related}

MRI super-resolution (SR) aims to recover high-resolution volumes from
anisotropic acquisitions that are common in clinical practice. By
enhancing inter-plane fidelity, SR can support more reliable
volumetric assessment, multiplanar visualization, and longitudinal
treatment planning. In retrospective brain MRI cohorts, the target
modality, such as T2-weighted (T2w) MRI, is often available only as a
single anisotropic volume per subject. These cohorts also exhibit
substantial protocol heterogeneity across scanners, sites, and
disease-specific imaging protocols. We summarize these constraints as
two defining properties of our setting.
\textbf{(C1)~No HR T2w.} The cohort contains no isotropic
high-resolution T2w, so paired low-resolution and high-resolution T2w
supervision cannot be assembled.
\textbf{(C2)~Unknown degradation.} Because no isotropic T2w was
acquired, the operator mapping a hypothetical HR T2w volume to the
observed anisotropic T2w is not measured and may vary across protocols.
% Beyond these data constraints, lesion-sensitive neuro-oncology further
% requires anatomical reliability. The reconstruction should preserve
% subject-specific anatomy and remain coherent across slices and planes,
% since fabricated structures or slice-wise inconsistencies can distort
% lesion boundaries and volumetric assessment~\cite{cohen2018distribution,
% iglesias2023synthsr}.

Existing brain MRI SR methods generally follow three complementary
paradigms. Model-based reconstruction treats SR as an inverse problem
by explicitly modeling the acquisition process. Learning-based SR uses
neural networks to learn a mapping from low-resolution inputs to
high-resolution outputs. Cross-modality (\ie cross-contrast) and reference-guided methods
use a companion modality to provide high-resolution anatomical
information when the target modality lacks an isotropic reference. We
review these paradigms with respect to C1, C2, and anatomical
reliability.

\subsection{Model-based Reconstruction}

Classical model-based reconstruction formulates SR as an inverse
problem, assuming that the observed low-resolution slices are generated
from a latent high-resolution volume through a predefined physical
forward operator~\cite{rousseau2006registration,gholipour2010robust}.
This operator can model slice selection, point-spread blur,
downsampling, and patients' motion when present. Because reconstruction is driven
by an explicit acquisition model, these methods do not require
high-resolution training data in the target modality and can therefore
operate under C1. Their performance, however, depends on how accurately
the assumed operator matches the true acquisition process. In
retrospective clinical cohorts, acquisition parameters often vary
across scanner hardware, pulse sequences, slice spacing, and
disease-specific protocols, making it difficult to characterize a
protocol-specific forward model. Thus, model-based reconstruction is
most suitable when the degradation process is known, measured, or
tightly bounded, whereas C2 remains a major limitation.

\subsection{Learning-based Super-Resolution}

Learning-based SR replaces the hand-designed forward operator with a
mapping learned from data. A major line of work trains this mapping
using paired low-resolution and high-resolution examples in the same
modality. These methods have achieved strong performance with
architectures such as dense 3D convolutional networks
\cite{chen2018dcsrn}, multiscale networks
\cite{pham2019multiscale,lyu2020multi}, implicit neural
representations for arbitrary-scale reconstruction
\cite{wu2022arbitrary}, slice-profile transformation
\cite{lin2023high}, adaptive scaling strategies
\cite{kim2024adaptive}, and GAN-based SR
\cite{wang2020enhancedgan}. When paired HR targets are available, the
network can learn the degradation implicitly from training data and
avoid specifying an explicit forward model. This partially addresses
C2, but it depends on the availability of isotropic HR T2w targets,
which violates C1 in our retrospective setting.

To reduce reliance on paired HR targets, self-supervised methods
construct supervision using its
high-resolution intra-plane content to guide recovery of the lower-resolution
inter-plane direction. SMORE uses internal
image statistics and anti-aliasing priors to improve inter-plane
resolution~\cite{zhao2020smore}. SIMPLE uses simultaneous multi-plane
self-supervision when orthogonal acquisitions are available
\cite{benisty2025simple}. ECLARE estimates the inter-plane slice profile from the input volume
and learns a self-supervised mapping between degraded and high-resolution
patches within the same scan
data~\cite{remedios2025eclare}. These approaches address C1 and attemp to address C2 by avoiding or estimating parts of the degradation
process. Their applicability nevertheless depends on information
available in the scan itself, such as reliable intra-plane statistics,
multiple acquisition planes, or an estimable slice profile.

Blind and perception-aware SR methods further address the unknown
degradation problem. Blind SR methods reduce the mismatch between
simulated downsampling and clinical degradation through unsupervised
domain transformation~\cite{zhou2022blind}. Generative-prior methods
instead use pretrained latent diffusion models or related architectures
to infer missing high-frequency information~\cite{wang2023inversesr,
zhao2025mamba}. These methods are valuable when the degradation model
is unavailable, but the recovered detail is largely constrained by the
population-level prior learned during training. In pathology-focused
neuro-oncology, this creates a potential risk: visually sharp details
may not be faithful to the patient's true T2w anatomy, especially for
lesions or postoperative changes that deviate from the training
distribution. Thus, perceptual quality alone is insufficient, motivating
patient-specific anatomical anchoring and explicit three-dimensional
coherence.

\subsection{Cross-Modality and Reference-Guided Synthesis}

When HR data are unavailable in the target modality, cross-modality
methods use another MR contrast from the same subject to provide
structural information. Multi-contrast synthesis methods learn
modality-invariant or modality-shared anatomical representations from
co-registered contrasts~\cite{chartsias2017multimodal}. Conditional
GANs and multi-stream translators, including mustGAN, can synthesize
missing or corrupted MR contrasts from available source images
\cite{dar2019image,yurt2021mustgan}. These methods demonstrate that a
companion modality can provide useful anatomical cues, but they
mainly optimize appearance fidelity. A synthesized T2w-like
image may be visually plausible while still deviating from
the patient's acquired T2w contrast.

Anatomical reliability is particularly important in lesion-sensitive
applications. Structure-aware translation methods, such as edge-aware
GANs, introduce boundary constraints to improve structural fidelity
during cross-modality synthesis~\cite{yu2019ea}. Unpaired volumetric
harmonization and latent diffusion models can further address
site-specific appearance differences through distribution-level
alignment~\cite{wu2026unpaired}.
These methods demonstrate the value of anatomical constraints, but
appearance-driven objectives may still prioritize stylistic consistency over
lesion-faithful preservation of patient-specific T2w information.

% Our approach is most closely related to reference-guided SR, where an
% external high-quality image provides fine-scale structural information
% for reconstruction~\cite{li2025cross}. Existing reference-guided
% formulations typically assume that the reference image provides
% high-resolution information compatible with the target reconstruction.
% In our setting, the available reference is cross-modality (\ie cross-contrast) rather than
% same-modality. We therefore use isotropic T1c as a high-resolution
% anatomical anchor while retaining the acquired anisotropic T2w as the
% patient-specific contrast source. This design addresses C1 without
% requiring same-modality isotropic HR T2w supervision, avoids the need
% for an explicit degradation operator under C2, and promotes anatomical
% reliability through projection-based fusion of three orthogonal T2w
% estimates.

Our approach is most closely related to reference-guided SR, where an
external high-quality image provides fine-scale structural information
for reconstruction~\cite{li2025cross}. Existing reference-guided
formulations typically assume that the reference image is compatible
with the target contrast or is drawn from the same imaging domain. In
our setting, however, the available reference is a different MRI
contrast. We therefore use isotropic T1c as a structural anchor rather
than as a substitute for T2w, while preserving the acquired anisotropic
T2w as the patient-specific contrast source. This setting motivates a
framework that can exploit cross-contrast anatomical guidance without
requiring isotropic HR T2w supervision or an explicit degradation
operator.

% =====================================================================
\section{Proposed Method}
\label{sec:method}
% =====================================================================

% \methodname is motivated by the fact that the acquisition plane of a
% clinical T2w scan retains high intra-plane resolution. On this plane, T2w
% contrast can be learned from the co-registered isotropic T1c, and the
% resulting mapping can be transferred to non-axial planes if the
% generator is prevented from encoding the slicing axis. The method
% therefore converts the absence of inter-plane T2w supervision into
% three orthogonal, axis-specific estimates of the same target volume,
% which are subsequently reconciled by a 3D fusion objective.
% Fig.~\ref{fig:overview} summarizes the two-stage framework.
% Stage~1 trains a view-independent patched generator (\vipgan) on axial
% T1c$\to$T2w pairs and applies the frozen generator along all three axes. Stage~2 fuses the three generated volumes using
% projection-based optimization, thereby imposing the 3D coherence that
% is not available from a slice-wise generator alone. Throughout both
% stages, all volumes remain on their native voxel grids where no resampling
% or interpolation is applied. This restriction is essential because
% resampling operators are axis-dependent and can introduce orientation
% specific artefacts that would compromise view transfer.

\methodname is built on the observation that the acquisition plane of a
clinical T2w scan retains high intra-plane resolution, while the
co-registered T1c provides isotropic anatomical structure. The method
uses this complementary information through a two-stage framework. In
Stage~1, a view-independent patched generator, \vipgan, is proposed and trained on
axial T1c$\to$T2w pairs and then applied unchanged along all three
anatomical axes. This converts the missing inter-plane supervision
into three orthogonal estimates of the same target T2w volume. In
Stage~2, these estimates are fused through projection-based
optimization, which imposes 3D coherence beyond what a slice-wise
generator can provide. Fig.~\ref{fig:overview} summarizes the proposed
framework.
Throughout both stages, all volumes remain on their native voxel grids,
and no resampling or interpolation is applied. This restriction is
important because resampling operators can be axis-dependent and may
introduce orientation-specific artifacts that compromise view transfer.

\subsection{Problem Formulation}
\label{sec:formulation}

Let $X\!\in\!\mathbb{R}^{H\times W\times D}$ denote a co-registered,
isotropic T1c volume and $Y\!\in\!\mathbb{R}^{H\times W\times D}$ the
patient's own anisotropic T2w volume, observed at full resolution along
the acquisition axis (axial, last dim) and at degraded inter-plane
resolution along the other two axes. Let $M\!\in\!\{0,1\}^{H\times W\times D}$
be the brain-tissue mask. Our goal is to estimate
\begin{equation}
    \widehat{Y}\!\in\!\mathbb{R}^{H\times W\times D} : \quad
    \widehat{Y}\big|_{M=1} \approx Y^{\text{iso}}_{M=1},
    \label{eq:goal}
\end{equation}
where $Y^{\text{iso}}$ is the (unobserved) isotropic T2w of the same
subject, with the constraint that no voxel of $Y^{\text{iso}}$ is
required at training or inference. The estimate $\widehat{Y}$ should
preserve patient-specific anatomy revealed by $X$ and remain consistent
with the contrast statistics of $Y$ along the high-resolution axis.

\subsection{Preprocessing and Patching Mechanism}
\label{sec:patch}

Patching is the mechanism that separates input standardization from
image resampling. Conventional SR pipelines often standardize input
dimensions by resampling the volume to a fixed resolution
~\cite{chen2018dcsrn,lyu2020multi}. In the present setting, this
operation is undesirable because the interpolation kernel acts along
different physical axes for axial, coronal and sagittal slices. The
resulting high-frequency artefacts can therefore become orientation
specific cues, encouraging the generator to learn a view-dependent
mapping. We instead use a deterministic, shape-preserving patching
scheme that changes only how each slice is partitioned, without
altering the voxel grid or applying interpolation.

\subsubsection{Intensity normalisation}
Foreground intensities are clipped to the $[0.5,99.5]$th percentile
range, z-score normalized within the brain mask, and shifted by a small positive value so
the foreground is strictly positive.

\subsubsection{Shape-preserving patching}
Given a 2D slice $S\in\mathbb{R}^{H\times W}$ extracted from a chosen
volume axis, we partition $S$ into $P{\times}P$ patches with $P\!=\!128$.
The number of patches along each spatial dimension is
\begin{equation}
    n_h = \left\lceil \frac{H}{P} \right\rceil,\qquad
    n_w = \left\lceil \frac{W}{P} \right\rceil,
    \label{eq:patchnum}
\end{equation}
and the strides are
\begin{equation}
    s_h = \frac{H-P}{n_h-1},\qquad s_w = \frac{W-P}{n_w-1},
    \label{eq:stride}
\end{equation}
so the patches cover the slice. The strides in~\eqref{eq:stride}
produce slightly overlapping patches whenever $H$ or $W$ is not a
multiple of $P$, and the overlaps are averaged at recombination, with
the foreground mask admitting only valid voxels. We set $P\!=\!128$, which
on BraTS volumes ($240{\times}240{\times}155$) yields $n_h\!=\!n_w\!=\!2$
patches per slice, balancing receptive field and batch
parallelism. 

Patching provides axial, coronal and sagittal slices the same fixed
$P{\times}P$ input format without changing the voxel grid. During
training, each T1c tile is paired with its T2w target. 
During inference,
the generator receives only the T1c tile and mask, with no explicit axis
label. Together with deepest-skip removal (Sec.~\ref{sec:vipgan}), this
encourages a patch-local cross-contrast mapping that can be reused across
all three views efficiently. 

\subsection{View-Independent Patched GAN (\vipgan)}
\label{sec:vipgan}

We  propose \vipgan as a 2D conditional GAN operating on the patches defined in
Sec.~\ref{sec:patch}. The architecture is designed to reconstruct
T2w-like contrast from T1c at the patch level while suppressing
view-specific representations and preserving the brain-mask boundary.

\subsubsection{Anatomy-robust generator}
The generator $G_\theta$ is a modified U-Net with four encoder blocks
$E_1\!-\!E_4$, a bottleneck $B$, and four decoder blocks $D_1\!-\!D_4$.
It differs from a standard U-Net in three respects. Each input patch
$x\in\mathbb{R}^{P\times P}$ is concatenated with its patch-local brain
mask $m\in\{0,1\}^{P\times P}$ to form a two-channel input
$\tilde{x}=[x,m]$. The skip connections do not carry full-resolution
encoder features. Instead, each encoder feature $F^{(\text{enc})}_l$ is
downsampled and then upsampled before being concatenated at the
corresponding decoder block,
\begin{equation}
    F^{(\text{dec})}_l \leftarrow \mathrm{Concat}\!\left(
    F^{(\text{dec})}_l,\,\mathrm{Up}_2\!\big(\mathrm{Down}_2(F^{(\text{enc})}_l)\big)
    \right).
    \label{eq:skip}
\end{equation}
The deepest skip connection between $E_4$ and $D_1$ is removed, so the bottleneck $B$ still receives the full $E_4$ output
while $D_1$ up-samples without concatenating $E_4$. The generator emits
a single-channel patch
$\hat{y}=G_\theta(\tilde{x})\in\mathbb{R}^{P\times P}$ of predicted T2w
intensities with the same spatial extent as the input.

These modifications serve the common objective of reducing sensitivity
to the slicing plane. The mask channel is particularly important for
non-axial slices near the superior and inferior brain boundaries, where
large background regions can otherwise dominate the adversarial signal.
Providing the mask separates anatomical tissue from air and prevents
the discriminator from using the brain--background boundary as a
shortcut cue. The modified skip connections address two additional
failure modes in cross-contrast, inter-plane synthesis. Full-resolution
skips can allow direct copying of T1c edges into the output, yielding a
contrast-shifted T1c image rather than a T2w-like prediction. Coarsened
skips reduce this shortcut and require the decoder to learn T2w
appearance from lower-resolution anatomical context. The same constraint
also supports view transfer: fine-scale T1c detail that is paired with
high-resolution T2w structure on axial training slices may not have an
equivalent target on anisotropic coronal or sagittal views. Removing the
deepest skip further forces global anatomy to pass through the shared
bottleneck representation, which is independent of orientation. The
effect of this architectural choice is illustrated qualitatively in
Fig.~\ref{fig:results_views}.

\subsubsection{Anatomy-aware discriminator}
Instead of evaluating a
T2w patch in isolation, the proposed  discriminator $D_\phi$ is conditional, which 
evaluates the paired T1c and candidate T2w patches and
estimates whether the candidate T2w is both realistic and anatomically
consistent with the co-registered T1c. This conditioning makes the
adversarial signal patient-specific.

The T1c and candidate T2w patches are passed through separate frozen
ImageNet-pretrained ResNet-50 feature extractors. Their features are
then fused by a shared $1{\times}1$ convolution and a lightweight
fully connected head that produces the real or fake logit. Freezing the
backbone and training only the fusion head limits overfitting to
low-level texture cues and encourages the discriminator to focus on
cross-contrast structural consistency, which also stabilizes
adversarial training.

\subsubsection{Adversarial training objective}
For an axial patch pair $(\tilde{x},y_{\text{real}})$ with T1c
conditioning channel $x$, the conditional discriminator is trained with
binary cross-entropy
\begin{equation}
    \mathcal{L}_D = \tfrac{1}{2}\!\left(
    \mathrm{BCE}(D_\phi(x, y_{\text{real}}), 1) +
    \mathrm{BCE}(D_\phi(x, G_\theta(\tilde{x})), 0)
    \right).
    \label{eq:discloss}
\end{equation}
The generator combines an adversarial term and a patch-level $L_1$
reconstruction term,
\begin{equation}
    \mathcal{L}_G = \mathcal{L}_{\text{adv}} +
                    \lambda_{\text{rec}}\,\mathcal{L}_{\text{rec}},
    \label{eq:genloss}
\end{equation}
\begin{equation}
    \mathcal{L}_{\text{adv}} =
    \mathrm{BCE}\!\left(D_\phi(x, G_\theta(\tilde{x})), 1\right),
    \label{eq:advloss}
\end{equation}
\begin{equation}
    \mathcal{L}_{\text{rec}} =
    \big\| (G_\theta(\tilde{x}) - y_{\text{real}}) \odot m \big\|_1,
    \label{eq:recloss}
\end{equation}
with $\lambda_{\text{rec}}\!=\!100$ chosen by validation. The mask
$m$ in~\eqref{eq:recloss} ensures that reconstruction error is
accumulated only over brain-tissue pixels, consistent with the
intensity-sentinel preprocessing in Sec.~\ref{sec:patch}. Training is
performed exclusively on \emph{axial} T1c$\to$T2w pairs.

\subsection{View Transfer}
\label{sec:transfer}
The generator $G_\theta$ trained on axial slices is reused directly on
coronal and sagittal views. Across orientations, only the slicing
and reconstruction geometry changes. The network weights, architecture,
and patch operator remain fixed. This view transfer is enabled by the
fixed square patch size $P=128$. Although axial slices are
$240\times240$ and coronal or sagittal slices are $240\times155$, all
views are decomposed into $128\times128$ patches before entering the
generator. As a result, orientation-specific intra-plane aspect ratios are
handled by the patching and reconstruction procedure rather than by the
network itself.

\begin{figure}[!t]
    \centering
    \includegraphics[width=0.74\textwidth]{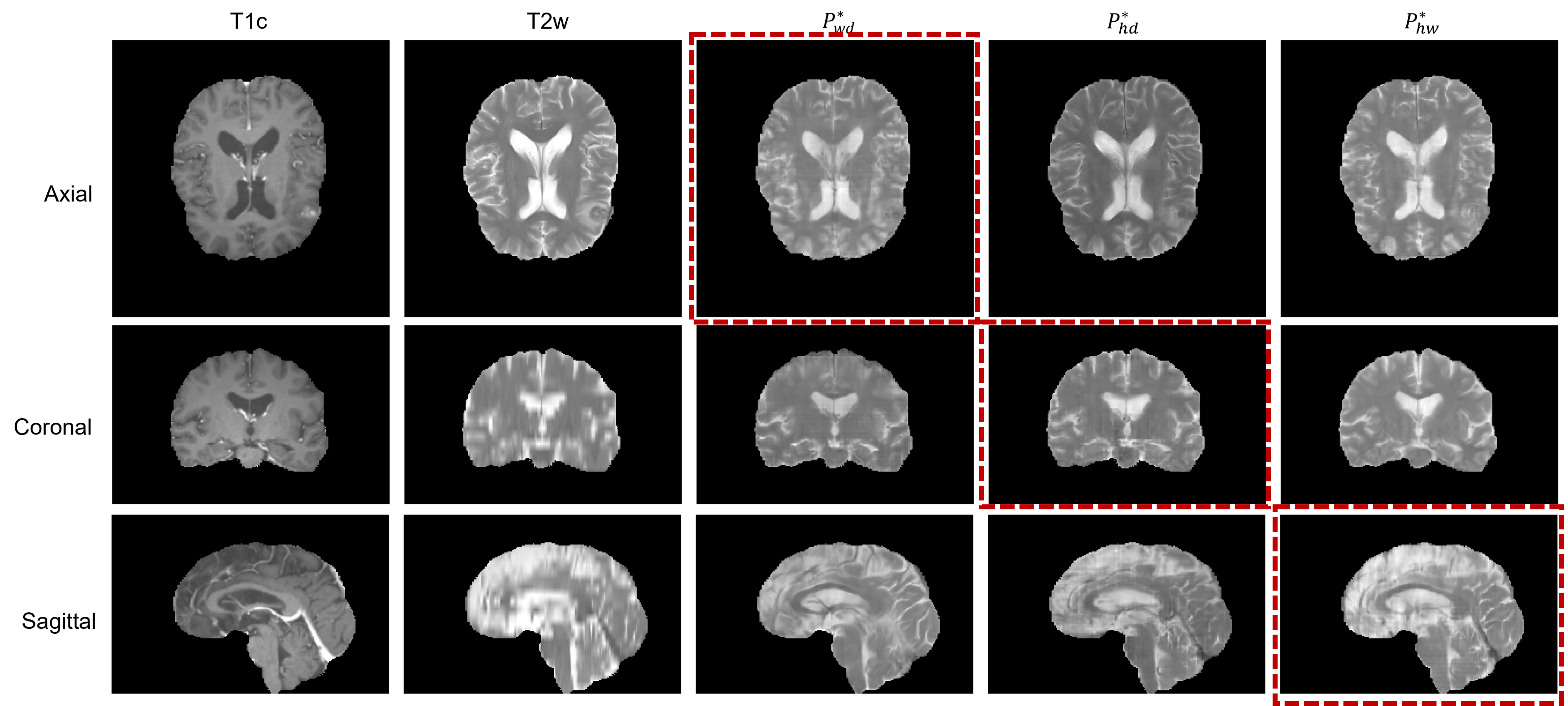}
    \caption{View-dependent outputs of the trained \vipgan. Top row:
    patient's co-registered T1c and the original T2w. Bottom row:
    the three view-specific reconstructions $P^*_{hw}, P^*_{hd},
    P^*_{wd}$ obtained by applying the same trained generator weights
    to axial, coronal and sagittal views in stage 1. Each $P^*_v$ is sharpest
    in its own slicing plane (boxes), highlighted by red boxes.}
    \label{fig:results_views}
\end{figure}

For each anatomical axis, the test T1c volume is decomposed into 2D
planes, patched as described in Sec.~\ref{sec:patch}, and processed by
$G_\theta$ patch by patch. The patch predictions are then reconstructed
into a 3D synthetic T2w volume. The coronal and sagittal outputs are
reoriented to the canonical axial grid, yielding three view-specific
volumes, $P^*_{hw}$, $P^**{hd}$, and $P^**{wd}$, in a shared coordinate
frame. Each volume is sharpest along the axis from which it was
generated and smoother in the orthogonal planes
(Fig.~\ref{fig:results_views}). The fusion stage exploits this
complementary anisotropy by treating inter-plane disagreement as an
indicator of view-specific uncertainty and resolving it through the
projection-based objective in Sec.~\ref{sec:fusion}.

\subsection{Projection-Based Optimization for 3D Fusion}
\label{sec:fusion}

Let $P\in\mathbb{R}^{H\times W\times D}$ be the unknown fused volume.
We seek
\begin{equation}
    P^{\star} = \arg\min_{P}\; \mathcal{L}(P)
    \quad\text{s.t.}\quad P = P\odot M,
    \label{eq:minP}
\end{equation}
where $M$ is the brain mask and $\mathcal{L}$ is one of the candidate
objectives below. Optimization is performed by Adam with learning
rate $10^{-3}$, maximum $10{,}000$ iterations, and a relative-tolerance
early-stopping criterion of $10^{-4}$.

\subsubsection{Self-consistency term.}
We define the slice-mean of $P$ along axis $a\!\in\!\{0,1,2\}$ as
$\overline{P}_a$ broadcast back over that axis, and write the
inter-plane variance penalty as
\begin{equation}
    \mathcal{L}_{\text{proj}}(P) =
    \sum_{a\in\{0,1,2\}} \big\| (P-\overline{P}_a)\odot M \big\|_F^2.
    \label{eq:Lproj}
\end{equation}
A perfectly self-consistent volume satisfies $P=\overline{P}_a$ along
every axis, which holds (within $M$) if and only if the volume is
\emph{constant} along that axis. Equation~\eqref{eq:Lproj} therefore
penalizes deviation from constancy along every axis simultaneously,
acting as a strong 3D-coherence prior in the absence of any HR ground
truth.

\subsubsection{View-fidelity term.}
We define the per-axis fidelity penalty as
\begin{equation}
    \mathcal{L}_{\text{dir}}(P) =
    \sum_{a\in\{0,1,2\}} \big\| (P - P^*_a)\odot M \big\|_F^2,
    \label{eq:Ldir}
\end{equation}
where $\{P^*_0, P^*_1, P^*_2\} = \{P^*_{hd}, P^*_{wd}, P^*_{hw}\}$ are
the three view-specific generator outputs from Sec.~\ref{sec:transfer}.
Equation~\eqref{eq:Ldir} ensures that the fused volume retains the
patient-specific anatomy each view's generator extracted from T1c.

\subsubsection{Auxiliary T2w-anchor term.}
An optional anchor against the patient's anisotropic T2w,
$\mathcal{L}_{\text{aux}}(P) = \| (P - Y)\odot M \|_F^2$
(\,$Y$ broadcast to isotropic along the high-resolution axis), is used
only in the ablation variant v3 below to test whether anchoring to the
acquired T2w adds signal beyond the T1c-derived views.

\subsubsection{The \methodname objective and its ablations}
The two penalties are complementary. $\mathcal{L}_{\text{dir}}$
preserves the patient-specific anatomy recovered by each T1c-derived
view, whereas $\mathcal{L}_{\text{proj}}$ regularizes inter-plane
disagreement toward a coherent 3D volume.  \methodname
therefore optimizes the weighted sum
\begin{equation}
    \mathcal{L}_{\star}(P) =
    \alpha\,\mathcal{L}_{\text{proj}}(P) + \beta\,\mathcal{L}_{\text{dir}}(P),
    \label{eq:s4}
\end{equation}
with weights $\alpha$ and $\beta$ chosen on a validation split. In experiments we set $\alpha\!=\!0.7,\ \beta\!=\!0.3$.
 Three reduced variants
attribute the contribution of each term and are reported as ablations
in Sec.~\ref{sec:results}. \textbf{v1} keeps only
$\mathcal{L}_{\text{dir}}$ and isolates what self-consistency adds.
\textbf{v2} drops optimization and simply averages the three view
volumes, the closed-form minimiser of $\mathcal{L}_{\text{dir}}$.
\textbf{v3} adds the auxiliary anchor,
$\mathcal{L}_{v3}\!=\!\alpha'\mathcal{L}_{\text{proj}}+\beta'(
\mathcal{L}_{\text{dir}}+\gamma\,\mathcal{L}_{\text{aux}})$ with
$\alpha'\!=\!0.9,\beta'\!=\!0.3,\gamma\!=\!1$, and tests whether
anchoring to the acquired anisotropic T2w adds signal beyond the
T1c-derived views.

\subsection{Algorithm and Practical Deployment}
\label{sec:algorithm}

The two stages of \methodname (Fig.~\ref{fig:overview}) have distinct
computational roles. \textbf{Stage~1} is a one-time cohort-level
training step in which \vipgan is optimised on axial T1c$\to$T2w pairs
from a representative cohort; the resulting checkpoint $G_\theta$ is
then frozen. \textbf{Stage~2} is the per-subject deployment step: the
frozen generator is applied along the three axes, and the resulting
volumes are fused by projection optimization. Stage~2 optimizes only the $HWD$ voxels of the output
volume through $O(HWD)$ reductions. 
% In our implementation it converges
% in $800$--$1{,}500$ iterations, taking approximately $30$\,s on one A100,
% with generator inference adding roughly $15$\,s.

This separation makes the expensive training stage amortized across all
future subjects. Applying the framework to a new subject or cohort
requires frozen-generator inference plus the parameter-free fusion step,
with no additional HR T2w data, degradation model, or retraining. The
BraTS-GLI experiments in Sec.~\ref{sec:results} follow this protocol.

\begin{algorithm}[!t]
\caption{\methodname}
\label{alg:vipsr}
\begin{algorithmic}[1]
\REQUIRE T1c volume $X$, anisotropic T2w volume $Y$, brain mask $M$
         (all in $\mathbb{R}^{H\times W\times D}$); weights $\alpha,\beta$.
\ENSURE  isotropic T2w estimate $\widehat{Y}$.
\STATE \textbf{Stage 1 (one-time, per cohort).} Train \vipgan on axial
       T1c$\to$T2w patch pairs; freeze the generator $G_\theta$.
\STATE \textbf{Stage 2 below is run per subject. Reuses $G_\theta$ unchanged.}
\STATE Pre-process $X, Y$ following Sec.~\ref{sec:patch}.
\FOR{each view $v \in \{hw, hd, wd\}$}
    \STATE Slice $X$ along axis $v$; patch via
           \eqref{eq:patchnum}--\eqref{eq:stride}.
    \STATE Apply $G_\theta$ per patch (with mask channel); recombine
           into $P^*_v$.
\ENDFOR
\STATE Initialise $P\leftarrow \tfrac{1}{3}(P^*_{hw}+P^*_{hd}+P^*_{wd})$.
\WHILE{not converged ($\Delta\mathcal{L}_{\star}/\mathcal{L}_{\star}>10^{-4}$)}
    \STATE $P \leftarrow P - \eta\,\nabla_P\mathcal{L}_{\star}(P)$ via Adam,
           using \eqref{eq:Lproj}--\eqref{eq:Ldir}.
    \STATE Project $P\leftarrow P\odot M$ (mask constraint).
\ENDWHILE
\RETURN $\widehat{Y}\leftarrow P$.
\end{algorithmic}
\end{algorithm}

% =====================================================================
\section{Experiments}
\label{sec:experiments}
% =====================================================================

\subsection{Datasets and Cross-Cohort Protocol}

We evaluate on two sub-cohorts from the Brain Tumor Segmentation
(BraTS)~2024 challenge~\cite{de20242024}. \textbf{BraTS-MET}
(brain metastasis, selected 220 subjects with anisotropic T2w) serves as the training cohort. We use
a random subject-level train/test split (147/73), and train the
view-independent generator \vipgan (Stage~1) only on the 147 training
subjects. \textbf{BraTS-GLI} (post-treatment adult glioma, 807 subjects)
is reserved for cross-cohort generalization. The BraTS-MET-trained
\vipgan checkpoint is applied to BraTS-GLI without retraining or
fine-tuning. Only the parameter-free Stage~2 projection optimization is
run per subject. The BraTS-GLI results therefore measure zero-shot
transfer and are reported on a 202-subject test split.

Each cohort carries voxel-wise tumour annotations, used only by the
downstream segmentation evaluation.
BraTS-MET provides three labels: enhancing tumour, tumour core and
peritumoural oedema. BraTS-GLI provides the four BraTS~2024
post-treatment labels~\cite{de20242024}. \emph{Enhancing tissue} (ET)
denotes contrast-enhancing tumour with blood--brain barrier disruption
on contrast-enhanced T1-weighted (T1c) magnetic resonance imaging
(MRI). \emph{Non-enhancing tumour core} (NETC) denotes the central
non-enhancing region, typically necrosis or cyst. \emph{Surrounding
non-enhancing fluid-attenuated inversion recovery (FLAIR)
hyperintensity} (SNFH) denotes oedema, infiltrating tumour, gliosis and
treatment-related change. \emph{Resection cavity} (RC) denotes the
surgical cavity after tumour removal. NETC and ET are small focal
labels, whereas SNFH is spatially dominant.

\subsection{Evaluation Strategy}
Paired isotropic high-resolution (HR) T2-weighted (T2w) ground truth is
unavailable.
Hence, conventional fidelity metrics against an HR target, such as
peak signal-to-noise ratio (PSNR), cannot quantify reconstruction
accuracy in this setting. We instead evaluate intensity fidelity in the
T2w domain, structural fidelity across modalities (\ie contrasts), and downstream
segmentation performance.

Before metric computation, all volumes are reoriented to the nearest
canonical right-anterior-superior (RAS) space and resampled to a common
grid. To reduce intensity-range bias across methods, we apply robust
within-mask clipping (0.5--99.5 percentile), within-mask normalization,
and zero-valued background masking. Unless otherwise specified, metrics
are computed within the brain mask $\Omega$ on the largest-mask-area
slice from each anatomical plane (axial, coronal and sagittal), and are
then aggregated across planes.

\textbf{(i)~Intensity fidelity (T2w domain).} The reference is the
patient's own acquired T2-weighted volume (T2w$_{\text{orig}}$).
Although anisotropic, it is the most faithful available reference for
T2w intensity statistics. The primary metric is intensity-domain
structural similarity
$\text{SSIM}_{\text{Int}}$~\cite{wang2004ssim}, computed after symmetric
$[0,1]$ normalization within the mask. We also report Laplacian-domain
root mean squared error (Lap-RMSE), structural similarity
($\text{SSIM}_{\text{Lap}}$), and high-frequency error norm (HFEN),
all referenced to T2w$_{\text{orig}}$.

\textbf{(ii)~Structural fidelity (cross-modality edges).} The reference
is the \emph{T1c} edge map, a different modality from the output. The
metric is Canny-Dice~\cite{canny1986}, defined as the Dice similarity
coefficient between the synthetic-T2w and T1c Canny edge maps.
Anatomical boundaries and lesion-adjacent structures remain consistent
across modalities, so a high score indicates inter-plane structure
consistent with the subject's anatomy, which the anisotropic acquisition
cannot resolve.

\textbf{(iii)~Downstream segmentation.} For each method (real T2w,
the ablations v1/v2/v3, and \methodname), we train an independent
nnU-Net model~\cite{isensee2021nnu} using the synthetic
or real T2w as the sole input channel and the cohort labels as targets.
The labels are never used during synthesis. BraTS-MET provides the
primary evaluation because it is the training cohort for \vipgan, whereas
BraTS-GLI measures generalization of the same frozen generator to an
unseen tumour type. Segmentation accuracy is also reported using the
Dice similarity coefficient (Dice).

\textbf{Metric definitions.}
Let $E_{\hat I}$ and $E_{\text{T1c}}$ be the binary Canny edge maps of
the synthetic T2w and the reference T1c within $\Omega$. The
cross-modality structural score is their Dice overlap,
\begin{equation}
\text{CannyDice}=\frac{2\,|E_{\hat I}\cap E_{\text{T1c}}|}
{|E_{\hat I}|+|E_{\text{T1c}}|}.
\label{eq:cannydice}
\end{equation}
For two images $a,b$ with local means $\mu_a,\mu_b$, variances
$\sigma_a^2,\sigma_b^2$ and covariance $\sigma_{ab}$, the structural
similarity index is
\begin{equation}
\text{SSIM}(a,b)=\frac{(2\mu_a\mu_b+c_1)(2\sigma_{ab}+c_2)}
{(\mu_a^2+\mu_b^2+c_1)(\sigma_a^2+\sigma_b^2+c_2)},
\label{eq:ssim}
\end{equation}
which we evaluate on intensity ($\text{SSIM}_{\text{Int}}$) and on the
Laplacian responses defined below ($\text{SSIM}_{\text{Lap}}$). To assess
high-frequency consistency, we compare the Laplacian responses of the
generated T2w and the reference T2w$_{\text{orig}}$ on each slice within
$\Omega$. Let $\nabla^2 I$ denote the discrete Laplacian. We consider
two responses: the signed Laplacian
$\mathcal{L}^{(2)}(I)=\nabla^2 I$, which retains the bipolar edge
structure (the paired positive and negative lobes and the zero-crossings
of an intensity transition), and its absolute counterpart
$\mathcal{L}^{(1)}(I)=|\nabla^2 I|$, which keeps only edge magnitude and
discards polarity. For each response $m\in\{1,2\}$, prediction and
reference maps are rescaled to $[0,1]$ using the dynamic range of the
reference map,

\begingroup
\small
\begin{equation}
\widehat{\mathcal{L}}^{(m)}(I)(x)=
\frac{\mathcal{L}^{(m)}(I)(x)-\ell^{(m)}_{\min}}
{\ell^{(m)}_{\max}-\ell^{(m)}_{\min}+\epsilon},
\quad x\in\Omega,
\label{eq:lap_norm}
\end{equation}
where $\ell^{(m)}_{\min}$ and $\ell^{(m)}_{\max}$ are the extrema of
$\mathcal{L}^{(m)}(I_{\text{ref}})$ over $\Omega$. On these rescaled
maps, we report range-normalized RMSE, HFEN and structural similarity,

\vspace{-0.4em}
\begin{equation}
\text{Lap-RMSE}^{(m)}=
\sqrt{\frac{1}{|\Omega|}\sum_{x\in\Omega}
\big(\widehat{\mathcal{L}}^{(m)}(I_{\text{pred}})(x)
-\widehat{\mathcal{L}}^{(m)}(I_{\text{ref}})(x)\big)^2},
\label{eq:lap_rmse}
\end{equation}

\vspace{-0.4em}
\begin{equation}
\text{HFEN}^{(m)}=
\frac{\big\|\widehat{\mathcal{L}}^{(m)}(I_{\text{pred}})
-\widehat{\mathcal{L}}^{(m)}(I_{\text{ref}})\big\|_{2}}
{\big\|\widehat{\mathcal{L}}^{(m)}(I_{\text{ref}})\big\|_{2}+\epsilon},
\label{eq:hfen}
\end{equation}

\vspace{-0.4em}
\begin{equation}
\text{SSIM}_{\text{Lap}}^{(m)}=
\text{SSIM}\big(\widehat{\mathcal{L}}^{(m)}(I_{\text{pred}}),\,
\widehat{\mathcal{L}}^{(m)}(I_{\text{ref}})\big).
\label{eq:ssim_lap}
\end{equation}
\endgroup

The subscripts L1 and L2 in
Tables~\ref{tab:phaseG_met}--\ref{tab:phaseG_gli} index these two
responses: L1 is the absolute Laplacian $|\nabla^2 I|$ and L2 the signed
Laplacian $\nabla^2 I$, with each metric otherwise computed identically.
The absolute response measures whether high-frequency structure is
present at a location, treating bright-to-dark and dark-to-bright
transitions alike. The signed response additionally requires transition
polarity to match the reference, and therefore penalizes fabricated
edges or local contrast inversions more strongly. The two responses
separate edge presence from edge correctness. Because the $[0,1]$
rescaling uses the reference range, Lap-RMSE is already
range-normalized.

\subsection{Implementation Details}
\vipgan was trained once on axial T1c$\to$T2w pairs from the 147
BraTS-MET training subjects until convergence. Convergence was defined
by the relative change in the training objective falling below a
predefined tolerance over consecutive monitoring intervals.
We use Adam with
$\beta_1\!=\!0.5,\beta_2\!=\!0.999$, learning rate
$2\!\times\!10^{-4}$ for both $G_\theta$ and $D_\phi$,
$\lambda_{\text{rec}}\!=\!100$, and a one-cycle linear schedule.
Patches are $P\!=\!128$. 
The fusion stage uses Adam with $\eta\!=\!10^{-3}$, maximum $10{,}000$
iterations and the early-stopping criterion in Sec.~\ref{sec:fusion}.
All code, including the patching utilities, training scripts, and
metric definitions, is released publicly (link withheld for
double-blind review).

\subsection{Computational Complexity}
At test time, \methodname consists of three feed-forward generator
passes followed by voxel-space fusion. With fixed patch size and fixed
network architecture, the generator inference cost scales linearly with
the voxel count $N=HWD$. The fusion stage directly optimizes the output
volume and uses only masked projections, per-axis reductions, and
elementwise differences. Each fusion update is therefore $O(N)$ and
requires no network backpropagation, giving an overall fusion
complexity of $O(N_{\mathrm{iter}}N)$. The only learned model stored at
inference is the $20.6$M-parameter generator.

For the selected diffusion-prior SR as the baseline, the computational profile differs. 
InverseSR
optimizes a compact latent representation, but each update requires
backpropagation through a large frozen latent diffusion model. In
contrast, \methodname performs a higher-dimensional voxel-space
optimization with inexpensive linear-time updates. Empirically, on the
same device (\ie with one single RTX5090), InverseSR requires more than ten hours for one case,
whereas the proposed Stage~2 fusion completes within seconds. Thus,
\methodname avoids repeated generative-prior backpropagation while
retaining linear scaling with image size during fusion.

\subsection{Choice of Baseline}

We benchmark against a single external method, InverseSR. This choice is
determined by the assumptions in Sec.~\ref{sec_related}.
Paired-supervision
approaches~\cite{chen2018dcsrn,pham2019multiscale,lyu2020multi,
lin2023sptsr,kim2024adaptive,luo2024target} require paired
high-resolution/low-resolution (HR/LR) T2w data for training, which are
unavailable in our setting (challenge~C1); such models therefore cannot
be trained on our cohort.
Self-supervised and blind SR
methods~\cite{zhao2020smore,benisty2024simple,remedios2025eclare,
zhu2017unpaired,xie2022parallelcyclegan} relax the paired-data
requirement but still assume a downsampling operator that is either
supplied or estimable from the data, typically from multi-orientation
T2w acquisitions or a 1D slice profile (challenge~C2). Neither of these
is available in our cohort. InverseSR is the only published method
with assumptions aligned to this setting: it performs SR by inverting a
generative prior learned from an external corpus, requires neither paired
HR/LR T2w nor an explicit degradation model, and operates on a single
anisotropic volume.

For this baseline we adopt the latent diffusion prior released by the
authors of InverseSR, which was pretrained on T1-weighted brain MRI. We
do not retrain the prior on T2w data, as no sufficiently large isotropic
T2w corpus is available for this purpose and our cohort provides no HR
T2w supervision. Using the authors' released T1-weighted prior thus
constitutes the reproducible configuration of
InverseSR under our data constraints.

% =====================================================================
\section{Results}
\label{sec:results}
% =====================================================================

\subsection{Image Quality}
% ---------------------------------------------------------------
% Phase-G image-quality metrics on BraTS-MET (220 cases, 3 planes)
% Best per column (excl. T2w_Original) is highlighted; proposed = VIPP-SR.
% ---------------------------------------------------------------
\begin{table}[t]
\centering
\footnotesize
\setlength{\tabcolsep}{2.8pt}
\renewcommand{\arraystretch}{1.3}
\caption{Image-quality metrics on BraTS-MET synthetic T2w (220 subjects,
3 orthogonal planes). Columns are grouped by evaluation purpose:
\emph{structural fidelity} measures cross-modality edge consistency,
scored against the T1c edge map (Canny edges); \emph{intensity fidelity}
measures agreement with the acquired T2w$_{\text{orig}}$. \textbf{Bold} =
best among non-oracle methods. T2w$_{\text{orig}}$ is the identity
baseline. $\uparrow$ higher is better, $\downarrow$ lower is better.}
\label{tab:phaseG_met}
\begin{tabular}{M*{7}{Y}}
\toprule
& \multicolumn{1}{c}{\textbf{Structural}}
& \multicolumn{6}{c}{\textbf{Intensity fidelity\ \ (T2w$_{\text{orig}}$) ref.}} \\
& \multicolumn{1}{c}{\textbf{(T1c ref.)}} & & & & & & \\
\cmidrule(lr){2-2}\cmidrule(lr){3-8}
\textbf{Method} & CannyDice $\uparrow$ & SSIM$_{\text{Int}}$ $\uparrow$ &
Lap$_{\text{RMSE,L1}}$ $\downarrow$ & SSIM$_{\text{Lap,L1}}$ $\uparrow$ &
HFEN$_{\text{L1}}$ $\downarrow$ &
SSIM$_{\text{Lap,L2}}$ $\uparrow$ & HFEN$_{\text{L2}}$ $\downarrow$ \\
\midrule
T2w$_{\text{orig}}$  & $0.275 \pm 0.049$ & $\equiv 1$ & $\equiv 0$ & $\equiv 1$ & $\equiv 0$ & $\equiv 1$ & $\equiv 0$ \\
\midrule
InverseSR      & $0.183 \pm 0.026$ & $0.397 \pm 0.095$ & $0.0856 \pm 0.014$ & $0.406 \pm 0.071$ & $0.918 \pm 0.079$ & $0.540 \pm 0.062$ & $0.109 \pm 0.019$ \\
\midrule
v1 (fidelity)        & \cellcolor{gray!20}\textbf{$0.464 \pm 0.058$} & $0.575 \pm 0.077$ & $0.0842 \pm 0.014$ & $0.467 \pm 0.074$ & $0.901 \pm 0.121$ & $0.582 \pm 0.071$ & $0.114 \pm 0.021$ \\
v2 (mean)            & $0.462 \pm 0.058$ & \cellcolor{gray!20}\textbf{$0.579 \pm 0.077$} & \cellcolor{gray!20}\textbf{$0.0840 \pm 0.014$} & $0.469 \pm 0.074$ & \cellcolor{gray!20}\textbf{$0.898 \pm 0.121$} & $0.583 \pm 0.070$ & $0.114 \pm 0.021$ \\
v3 ($+\mathcal{L}_{\text{aux}}$) & $0.442 \pm 0.055$ & $0.574 \pm 0.073$ & $0.0843 \pm 0.014$ & $0.477 \pm 0.074$ & $0.901 \pm 0.110$ & \cellcolor{gray!20}\textbf{$0.603 \pm 0.069$} & \cellcolor{gray!20}\textbf{$0.113 \pm 0.020$} \\
\textbf{\methodname} & $0.449 \pm 0.056$ & $0.578 \pm 0.074$ & \cellcolor{gray!20}\textbf{$0.0840 \pm 0.014$} & \cellcolor{gray!20}\textbf{$0.477 \pm 0.073$} & \cellcolor{gray!20}\textbf{$0.898 \pm 0.112$} & $0.599 \pm 0.069$ & $0.113 \pm 0.020$ \\
\bottomrule
\end{tabular}
\end{table}

% ---------------------------------------------------------------
% Phase-G image-quality metrics on BraTS-GLI (807 cases, 3 planes).
% Zero-shot cross-cohort transfer: every learned method uses the SAME
% BraTS-MET-trained checkpoint without any retraining or fine-tuning
% on BraTS-GLI. T2w_orig is the identity baseline (real T2w vs. itself)
% and is excluded from bold-best comparisons.
% ---------------------------------------------------------------
\begin{table}[t]
\centering
\footnotesize
\setlength{\tabcolsep}{2.8pt}
\renewcommand{\arraystretch}{1.3}
\caption{Image-quality metrics on BraTS-GLI synthetic T2w (807 subjects,
3 orthogonal planes). Bold indicates the best learned method;
$\uparrow$ higher is better and $\downarrow$ lower is better.}
\label{tab:phaseG_gli}
\begin{tabular}{M*{7}{Y}}
\toprule
& \multicolumn{1}{c}{\textbf{Structural}}
& \multicolumn{6}{c}{\textbf{Intensity fidelity\ \ (T2w$_{\text{orig}}$ ref.)}} \\
& \multicolumn{1}{c}{\textbf{(T1c ref.)}} & & & & & & \\
\cmidrule(lr){2-2}\cmidrule(lr){3-8}
\textbf{Method} & CannyDice $\uparrow$ & SSIM$_{\text{Int}}$ $\uparrow$ &
Lap$_{\text{RMSE,L1}}$ $\downarrow$ & SSIM$_{\text{Lap,L1}}$ $\uparrow$ &
HFEN$_{\text{L1}}$ $\downarrow$ &
SSIM$_{\text{Lap,L2}}$ $\uparrow$ & HFEN$_{\text{L2}}$ $\downarrow$ \\
\midrule
% \multicolumn{8}{@{}l}{\emph{Identity baseline (real T2w vs.\ itself; reference, not a synthesis result)}} \\
T2w$_{\text{orig}}$  & $0.352 \pm 0.044$ & $\equiv 1$ & $\equiv 0$ & $\equiv 1$ & $\equiv 0$ & $\equiv 1$ & $\equiv 0$ \\
\midrule
% \multicolumn{8}{@{}l}{\emph{Prior art (single-image inversion; T1-weighted prior, no T2w retraining)}} \\
InverseSR     & $0.238 \pm 0.032$ & $0.555 \pm 0.058$ & $0.0921 \pm 0.011$ & $0.407 \pm 0.055$ & $0.855 \pm 0.029$ & $0.556 \pm 0.051$ & $0.115 \pm 0.013$ \\
\midrule
% \multicolumn{8}{@{}l}{\emph{\methodname and ablations (BraTS-MET checkpoint, zero-shot on BraTS-GLI)}} \\
v1 (fidelity)        & $0.496 \pm 0.051$ & $0.643 \pm 0.061$ & $0.0893 \pm 0.012$ & $0.488 \pm 0.056$ & $0.841 \pm 0.070$ & $0.622 \pm 0.057$ & $0.119 \pm 0.018$ \\
v2 (mean)            & \cellcolor{gray!20}\textbf{$0.496 \pm 0.051$} & \cellcolor{gray!20}\textbf{$0.644 \pm 0.061$} & $0.0893 \pm 0.012$ & $0.488 \pm 0.056$ & $0.840 \pm 0.070$ & $0.622 \pm 0.057$ & $0.119 \pm 0.018$ \\
v3 ($+\mathcal{L}_{\text{aux}}$) & $0.479 \pm 0.049$ & $0.631 \pm 0.056$ & $0.0895 \pm 0.012$ & $0.494 \pm 0.057$ & $0.842 \pm 0.062$ & \cellcolor{gray!20}\textbf{$0.641 \pm 0.055$} & $0.117 \pm 0.018$ \\
\textbf{\methodname} & $0.486 \pm 0.049$ & $0.635 \pm 0.057$ & \cellcolor{gray!20}\textbf{$0.0890 \pm 0.012$} & \cellcolor{gray!20}\textbf{$0.496 \pm 0.056$} & \cellcolor{gray!20}\textbf{$0.838 \pm 0.062$} & $0.639 \pm 0.056$ & \cellcolor{gray!20}\textbf{$0.117 \pm 0.018$} \\
\bottomrule
\end{tabular}

\end{table}

Tables~\ref{tab:phaseG_met}--\ref{tab:phaseG_gli} report image quality
on both cohorts, with columns grouped by the two evaluation axes of
Sec.~\ref{sec:experiments}. The structural column uses the T1c edge map
as an external anatomical reference, whereas the intensity-fidelity
columns use the acquired T2w$_{\text{orig}}$ as the reference. For
BraTS-GLI (Table~\ref{tab:phaseG_gli}), every learned row is produced
by the same BraTS-MET-trained \vipgan checkpoint, applied without
retraining or fine-tuning. The T2w$_{\text{orig}}$ row is therefore an
identity reference for intensity-fidelity metrics; its perfect entries
are mathematical identities rather than competitive measurements.

\textbf{Structural fidelity (cross-modality edges).}
All four learned methods substantially outperform the
T2w$_{\text{orig}}$ identity row on Canny-Dice against T1c. The score
rises from $0.275$ to $0.442$--$0.464$ on BraTS-MET ($\sim$65\%
relative) and from $0.352$ to $0.479$--$0.496$ on BraTS-GLI
($\sim$40\% relative). The latter is obtained under zero-shot
cross-cohort transfer, with the BraTS-MET checkpoint applied to glioma subjects
without any retraining or fine-tuning. CannyDice is the only
image-quality column whose reference (T1c) is independent of the
acquired T2w, so it is the column on which T2w$_{\text{orig}}$
constitutes a meaningful comparator rather than an identity
tautology; on that column, every variant of \methodname dominates
T2w$_{\text{orig}}$ on both cohorts by a wide margin. The synthetic
volumes are therefore not noisy copies of the acquired T2w; they
agree with the subject's own T1c anatomy more closely than the
anisotropic T2w itself does, recovering the inter-plane structure
the acquisition discarded.

\textbf{Intensity fidelity (T2w domain).}
The intensity-fidelity columns are referenced to T2w$_{\text{orig}}$
itself, so the T2w$_{\text{orig}}$ row appears in Tables~\ref{tab:phaseG_met} and
\ref{tab:phaseG_gli} as a notational anchor rather than as a competing
method. Among the learned methods, the rows separate into two groups.
The mean and fidelity variants (v2, and v1 on BraTS-GLI) score
highest on the pointwise $\text{SSIM}_{\text{Int}}$, while \methodname
and v3 lead on the Laplacian-domain metrics that probe high-frequency
structure ($\text{SSIM}_{\text{Lap}}$ and HFEN). This is the expected
effect of self-consistency: penalizing inter-plane variance trades a
small amount of pointwise intensity agreement for a gain in 3D
structural coherence. \methodname and v3 track each other closely on
every metric and cohort, with neither consistently ahead. \methodname
leads on $\text{SSIM}_{\text{Lap,L1}}$, Lap-RMSE and
$\text{HFEN}_{\text{L1}}$, and v3 on $\text{SSIM}_{\text{Lap,L2}}$ and
$\text{HFEN}_{\text{L2}}$. Once self-consistency is present, the
auxiliary T2w anchor therefore adds little measurable signal. Because
the BraTS-GLI numbers are produced by the frozen BraTS-MET checkpoint,
the residual intensity gap to the acquired glioma T2w is bounded above
by the unmodelled cross-cohort domain shift; cohort-specific
retraining of \vipgan, which is intentionally not performed in this
paper so as to isolate the generalization behaviour, is the natural
route to closing it.

\subsection{Downstream Segmentation}
% ---------------------------------------------------------------
% Downstream segmentation Dice on BraTS-MET (held-out test, n=73)
% ---------------------------------------------------------------
\begin{table}[t]
\centering
% \footnotesize
\setlength{\tabcolsep}{4pt}
\renewcommand{\arraystretch}{1.25}
\caption{Downstream segmentation Dice on BraTS-MET held-out test set ($n{=}73$).
nnU-Net v2 (3d\_fullres, 250 epochs, fold 0). Same label set on disk for all
rows; the only difference is the T2w volume given as the network's single
input channel. Mean $\pm$ SEM. \textbf{Bold} = best (excl.\ real T2w upper bound).}
\label{tab:seg_met}
\resizebox{\columnwidth}{!}{%
\begin{tabular}{>{\raggedright\arraybackslash}p{0.24\columnwidth}*{4}{>{\centering\arraybackslash}p{0.15\columnwidth}}}
\toprule
\textbf{Method} & Mean-label & Tumor (1) & Tumor core (2) & Edema (3) \\
\midrule
T2w$_{\text{orig}}$  & $0.330$ & $0.308$ & $0.475$ & $0.280$ \\
\midrule
InverseSR & $0.025$ & $0.000$ & $0.083$ & $0.006$ \\
\midrule
v1 (fidelity)            & \cellcolor{gray!20}\textbf{$0.484$} & $0.492$ & $0.485$ & \cellcolor{gray!20}\textbf{$0.529$} \\
v2 (mean)                & $0.453$ & $0.501$ & $0.456$ & $0.514$ \\
v3 ($+\mathcal{L}_{\text{aux}}$) & $0.449$ & $0.470$ & $0.478$ & $0.502$ \\
\textbf{\methodname}   & $0.465$ & \cellcolor{gray!20}\textbf{$0.556$} & \cellcolor{gray!20}\textbf{$0.485$} & $0.515$ \\
\bottomrule
\end{tabular}%
}
\end{table}

% ---------------------------------------------------------------
% Downstream segmentation Dice on BraTS-GLI (held-out test, n=202).
% Zero-shot cross-cohort transfer: the VIPGAN generator is trained
% only on BraTS-MET and applied to BraTS-GLI without any retraining
% or fine-tuning. T2w_orig is the in-domain real-T2w upper bound and
% is excluded from the bold-best comparison among learned methods.
% Daggers (†) mark synthetic cells that exceed the real-T2w reference.
% ---------------------------------------------------------------
\begin{table}[t]
\centering
\small
\setlength{\tabcolsep}{3pt}
\renewcommand{\arraystretch}{1.25}
\caption{Downstream segmentation Dice on the BraTS-GLI held-out test set
($n{=}202$). Bold indicates the best learned method; $\dagger$ marks a
synthetic value exceeding the real-T2w reference.}
\label{tab:seg_gli}
\resizebox{\columnwidth}{!}{%
\begin{tabular}{>{\raggedright\arraybackslash}p{0.20\columnwidth}*{5}{>{\centering\arraybackslash}p{0.13\columnwidth}}}
\toprule
\textbf{Method} & Mean-label $\uparrow$ & NETC (1) $\uparrow$ & SNFH (2) $\uparrow$ & ET (3) $\uparrow$ & RC (4) $\uparrow$ \\
\midrule
% \multicolumn{6}{@{}l}{\emph{In-domain real-T2w reference (no synthesis; segmenter sees acquired data)}} \\
T2w$_{\text{orig}}$  & $0.473$ & $0.140$ & \cellcolor{gray!20}\textbf{$0.796$} & $0.278$ & $0.530$ \\
\midrule
% \multicolumn{6}{@{}l}{\emph{Prior art (single-image inversion; T1-weighted prior, no T2w retraining)}} \\
InverseSR & $0.066$ & $0.005$ & $0.075$ & $0.008$ & $0.175$ \\
\midrule
% \multicolumn{6}{@{}l}{\emph{\methodname and ablations (BraTS-MET checkpoint, zero-shot on BraTS-GLI)}} \\
v1 (fidelity)            & $0.552^{\dagger}$ & $0.371^{\dagger}$ & $0.733$ & \cellcolor{gray!20}\textbf{$0.515^{\dagger}$} & $0.530$ \\
v2 (mean)                & $0.552^{\dagger}$ & $0.372^{\dagger}$ & $0.735$ & $0.488^{\dagger}$ & $0.522$ \\
v3 ($+\mathcal{L}_{\text{aux}}$) & $0.549^{\dagger}$ & \cellcolor{gray!20}\textbf{$0.377^{\dagger}$} & $0.730$ & $0.468^{\dagger}$ & \cellcolor{gray!20}\textbf{$0.544^{\dagger}$} \\
\textbf{\methodname}   & \cellcolor{gray!20}\textbf{$0.563^{\dagger}$} & \cellcolor{gray!20}\textbf{$0.377^{\dagger}$} & $0.737$ & $0.488^{\dagger}$ & $0.535^{\dagger}$ \\
\bottomrule
\end{tabular}%
}

\end{table}

Tables~\ref{tab:seg_met}--\ref{tab:seg_gli} report nnU-Net Dice on the
held-out test sets. The primary result is BraTS-MET, the cohort
\vipgan was trained on, and BraTS-GLI is reported afterwards as a
cross-cohort generalization check. For each row, nnU-Net
 is trained independently with that
method's T2w volume as the sole input channel; labels and train/test
splits are identical across rows. Mean-label Dice averages the per-class
Dice over all lesion sub-classes, and the per-class columns report
class-wise means. In
Table~\ref{tab:seg_gli}, dagger-marked entries denote synthetic
results that exceed the real-T2w reference.

\textbf{BraTS-MET (primary).}
On the 73 held-out metastasis subjects (Table~\ref{tab:seg_met}), the
reconstructed T2w improves downstream segmentation over the acquired
anisotropic T2w: \methodname raises mean-label Dice from $0.330$ to
$0.465$ and tumour Dice from $0.308$ to $0.556$, showing that the
reconstructed T2w provides a more effective single-channel input for
downstream tumour segmentation. As these are single-fold nnU-Net
results without cross-validation, we treat the gains as indicative
rather than formally significant; they are nonetheless notable because
\methodname is trained without isotropic T2w ground truth or
segmentation supervision. Among the reconstructed variants, \methodname
achieves the best tumour Dice ($0.556$) and matches the best
tumour-core Dice ($0.485$), while v1 gives the highest mean-label Dice
($0.484$) and edema Dice ($0.529$). The large relative gain on edema (\ie from $0.280$ with the
acquired T2w to $0.515$--$0.529$ for the reconstructed T2w) is
consistent with its morphology in metastasis: peritumoural edema is a
thin, T2w-subtle FLAIR-hyperintense rim around the enhancing core
(cohort median $\approx 2.4$~mL, only $\approx 19\%$ of the total lesion
volume) that a single anisotropic T2w only partially resolves, so a
sharper inter-plane reconstruction recovers substantial boundary
detail. The same peritumoural signal behaves very differently in the
glioma cohort, as discussed next.

\textbf{BraTS-GLI (zero-shot cross-cohort generalization).}
Table~\ref{tab:seg_gli} evaluates the same BraTS-MET trained generator
on 202 held-out glioma subjects without retraining, fine-tuning, or
cohort-specific adaptation. The acquired T2w reference remains
strongest on the spatially dominant SNFH class ($0.796$ vs.\ $0.737$
for \methodname). This gap reflects the morphology of SNFH in glioma:
infiltrative glioblastoma produces a large, confluent SNFH region
(cohort median $\approx 50$~mL, $\approx 74\%$ of total lesion
volume) that is already well resolved on a single anisotropic T2w,
leaving little headroom for reconstruction to improve upon, whereas
the smaller, focal classes leave more to recover. Accordingly,
\methodname achieves the best mean-label Dice among learned methods,
improving over the acquired T2w reference and exceeding it on the
smaller lesion-related classes NETC, ET, and RC. In contrast,
InverseSR performs poorly in this setting.
These results imply that the reconstructed T2w supplies
useful lesion-specific detail even under cross-cohort transfer, highlighting the importance of the proposed patching
and projection-based design.

Across both cohorts, \methodname is consistently among the strongest
learned reconstructions and is the only candidate that explicitly
couples view-specific estimates through 3D self-consistency. We
therefore adopt it as the final method and retain v1, v2, and v3 as
ablations.

\subsection{Qualitative Structural Evaluation}

\begin{figure}[!t]
    \centering
    \includegraphics[width=0.7\textwidth]{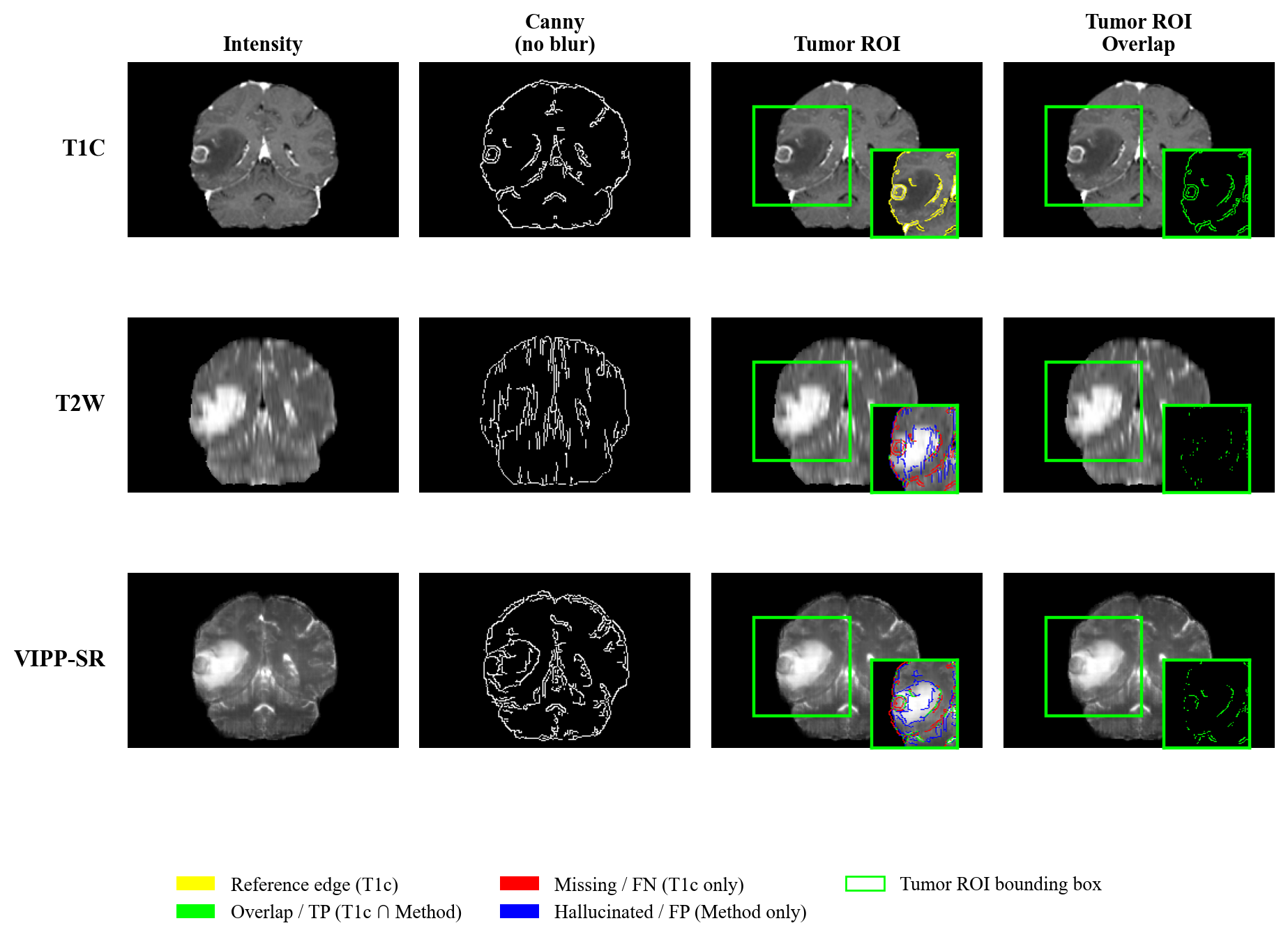}
    \caption{Structural edge comparison on a representative BraTS-GLI
    coronal reformat (case 00307, $y{=}84$). Rows show T1c, acquired T2w,
    and \methodname; columns show intensity, Canny edges, ROI edge
    agreement, and matched T1c edges. In ROI overlays, green, red and blue
    denote matched, missed and spurious edges, respectively.}
    \label{fig:phaseB_struct}
\end{figure}

Fig.~\ref{fig:phaseB_struct} illustrates the structural-fidelity
evaluation on a coronal reformat, where inter-plane degradation is
most visible. The acquired anisotropic T2w is blurred and
stair-stepped, leaving many missed T1c-defined edges along the tumour
boundary. In contrast, \methodname recovers the boundary with
predominantly matched edges and few missed or spurious edges, matching
the Canny-Dice gains in
Tables~\ref{tab:phaseG_met}--\ref{tab:phaseG_gli}.

\begin{table}[t]
\centering
\scriptsize
\setlength{\tabcolsep}{3pt}
\renewcommand{\arraystretch}{1.12}
\caption{Per-plane image-quality summary and Stage~2 ablation on
BraTS-MET. Values are mean $\pm$ SEM over evaluated cases within each
plane. NoOpt denotes direct view-specific generator output before
projection-based fusion. Higher Canny Dice and SSIM$_{\mathrm{Int}}$
indicate better agreement, whereas lower HFEN$_{L2}$ indicates lower
high-frequency error.}
\label{tab:plane_stage2}
\begin{tabular}{llccc}
\toprule
\textbf{Method} & \textbf{Plane} &
\textbf{Canny Dice} $\uparrow$ &
\textbf{SSIM$_{\mathrm{Int}}$} $\uparrow$ &
\textbf{HFEN$_{L2}$} $\downarrow$ \\
\midrule
NoOpt & Coronal  & $0.425{\pm}0.003$ & $0.516{\pm}0.005$ & $0.134{\pm}0.002$ \\
& Axial    & $0.418{\pm}0.003$ & $0.564{\pm}0.005$ & $0.129{\pm}0.001$ \\
& Sagittal & $0.404{\pm}0.004$ & $0.552{\pm}0.004$ & $0.111{\pm}0.001$ \\
\midrule
v1 (fidelity) & Coronal  & $0.471{\pm}0.003$ & $0.549{\pm}0.005$ & $0.123{\pm}0.001$ \\
& Axial    & $0.472{\pm}0.004$ & $0.572{\pm}0.005$ & $0.118{\pm}0.001$ \\
& Sagittal & $0.448{\pm}0.004$ & $0.604{\pm}0.005$ & $0.103{\pm}0.001$ \\
\midrule
v2 (mean) & Coronal  & $0.469{\pm}0.003$ & $0.554{\pm}0.005$ & $0.122{\pm}0.001$ \\
& Axial    & $0.472{\pm}0.004$ & $0.577{\pm}0.005$ & $0.117{\pm}0.001$ \\
& Sagittal & $0.446{\pm}0.004$ & $0.608{\pm}0.005$ & $0.103{\pm}0.001$ \\
\midrule
v3 ($+\mathcal{L}_{\mathrm{aux}}$) & Coronal  & $0.446{\pm}0.003$ & $0.551{\pm}0.005$ & $0.120{\pm}0.001$ \\
& Axial    & $0.447{\pm}0.004$ & $0.565{\pm}0.005$ & $0.117{\pm}0.001$ \\
& Sagittal & $0.434{\pm}0.004$ & $0.606{\pm}0.004$ & $0.101{\pm}0.001$ \\
\midrule
\textbf{\methodname} & Coronal  & $0.453{\pm}0.003$ & $0.554{\pm}0.005$ & $0.121{\pm}0.001$ \\
& Axial    & $0.455{\pm}0.004$ & $0.571{\pm}0.005$ & $0.117{\pm}0.001$ \\
& Sagittal & $0.439{\pm}0.004$ & $0.608{\pm}0.004$ & $0.102{\pm}0.001$ \\
\bottomrule
\end{tabular}
\end{table}

\subsection{Per-Plane Behaviour and Stage~2 Ablation}

Table~\ref{tab:plane_stage2} reports a per-plane comparison between
the direct generator output and the optimized reconstructions. NoOpt
denotes the view-specific generator outputs before Stage~2 fusion.
Across all three anatomical planes, the optimized variants improve
Canny-edge Dice and SSIM$_{\mathrm{Int}}$ and reduce HFEN$_{L2}$
relative to NoOpt. This consistent improvement demonstrates the value
of the projection-based optimization stage, showing that integrating
information from multiple views produces reconstructions with better
structural fidelity, intensity consistency, and preservation of
high-frequency details than direct patch-wise generation alone.

The per-plane results also show a stable residual anisotropy. Sagittal
slices achieve the strongest intensity and high-frequency agreement,
with the highest SSIM$_{\mathrm{Int}}$ and lowest HFEN$_{L2}$ across
the optimized variants. Coronal slices remain the most challenging for
these two intensity-based metrics, while Canny-edge Dice is generally
higher on axial and coronal planes than on sagittal planes. Importantly,
the performance gains from Stage~2 are observed across all planes,
indicating that the optimization process provides robust benefits
regardless of viewing direction. Because this plane-dependent pattern
is shared by \methodname and its ablations, the optimization stage
appears to preserve the strengths of the underlying view-specific
generators while improving inter-plane consistency. These findings
highlight the effectiveness of the fusion strategy in leveraging
complementary information from multiple anatomical planes to produce
more coherent and anatomically faithful isotropic reconstructions.

% =====================================================================
\section{Discussion}
\label{sec:discussion}
% =====================================================================

The BraTS-MET results validate the main contribution of \methodname. On the
cohort used to train the generator, the reconstructed T2w improves
mean-label segmentation Dice over the acquired anisotropic T2w with no
isotropic T2w ground truth and no segmentation supervision. It suggests
that cross-contrast anatomical guidance from T1c, combined with
projection-based 3D fusion, can produce a T2w representation that is at
least as useful as the acquired anisotropic T2w for downstream
single-channel segmentation. The gain is most relevant in settings
where small structures and inter-plane consistency matter, such as
metastatic lesion segmentation and other low-data neuro-oncology
applications.

The BraTS-GLI experiment provides a stricter zero-shot cross-cohort
assessment. The \vipgan generator is trained only on BraTS-MET and then
applied to BraTS-GLI without retraining, fine-tuning, or
cohort-specific adaptation. The acquired glioma T2w therefore serves as
an in-domain reference for the target cohort, while \methodname tests
how far the BraTS-MET trained reconstruction model can transfer. Under
this setting, \methodname remains the strongest learned reconstruction
and improves mean-label Dice over the acquired T2w reference. It also improves several smaller lesion-related classes,
including NETC, ET, and RC. The remaining gap on SNFH indicates that
glioma-specific T2w appearance, especially diffuse hyperintensity and
treatment-related changes, is not fully captured by a generator trained
on metastases. This behavior is consistent with a controlled
cross-cohort stress test.
Light cohort-specific fine-tuning of the generator can further improve the performance.

The per-plane analysis further clarifies where improvement is still
limited. Stage~2 fusion consistently improves over the direct generator
output across anatomical planes, confirming that the projection-based
optimization contributes beyond patch-wise synthesis alone. At the same
time, a residual plane-dependent pattern remains. Sagittal slices show
the strongest intensity and high-frequency agreement, whereas coronal
slices remain the most challenging on these metrics. Because this
pattern is shared by \methodname and its ablations, it is likely driven
by the view-specific generator outputs rather than by a single fusion
loss. Future improvements will therefore focus on stronger
view-independent generation, inter-plane feature learning, and possibly
view-aware discriminator designs, while retaining the 3D fusion
principle.

Several limitations remain. First, \methodname requires a high-quality
co-registered isotropic T1c volume. When such a cross-contrast anchor is
unavailable or is itself anisotropic, self-supervised SR methods that
operate only on the target scan may be more appropriate.  
Second, the downstream segmentation results support the
utility of the reconstructed T2w, but prospective clinical validation
and reader studies will be needed before the reconstruction can be
interpreted as a diagnostic replacement for acquired images.

% =====================================================================
\section{Conclusion}
\label{sec:conclusion}
% =====================================================================

We propose \methodname, a cross-contrast guided SR framework for
anisotropic T2w brain MRI. The method uses isotropic T1c as a
high-resolution anatomical anchor, preserves the acquired anisotropic
T2w as the patient-specific contrast source, and fuses three
view-specific estimates through a projection-based 3D optimization that
enforces anatomical consistency across them.
Without requiring isotropic T2w ground truth or segmentation
supervision, \methodname improves downstream segmentation over the
acquired anisotropic T2w on BraTS-MET and transfers without retraining
to BraTS-GLI. These results indicate that enforcing anatomical consistency across
view-specific reconstructions, together with view-independent
generation, provides a practical route toward retrospective T2w
super-resolution in heterogeneous neuro-oncology cohorts.

% =====================================================================
\section*{Acknowledgments}
% =====================================================================

\bibliographystyle{unsrt}
\bibliography{ref}

@article{teyateeti2020brain,
  title={Brain metastases resection cavity radio—surgery based on T2-weighted MRI: Technique assessment},
  author={Teyateeti, Achiraya and Brown, Paul D and Mahajan, Anita and Laack, Nadia N and Pollock, Bruce E},
  journal={Journal of neuro-oncology},
  volume={148},
  number={1},
  pages={89--95},
  year={2020},
  publisher={Springer}
}

@article{castellano2021advanced,
  title={Advanced imaging techniques for radiotherapy planning of gliomas},
  author={Castellano, Antonella and Bailo, Michele and Cicone, Francesco and Carideo, Luciano and Quartuccio, Natale and Mortini, Pietro and Falini, Andrea and Cascini, Giuseppe Lucio and Minniti, Giuseppe},
  journal={Cancers},
  volume={13},
  number={5},
  pages={1063},
  year={2021},
  publisher={MDPI}
}

@article{wu2022arbitrary,
  title={An arbitrary scale super-resolution approach for 3d mr images via implicit neural representation},
  author={Wu, Qing and Li, Yuwei and Sun, Yawen and Zhou, Yan and Wei, Hongjiang and Yu, Jingyi and Zhang, Yuyao},
  journal={IEEE Journal of Biomedical and Health Informatics},
  volume={27},
  number={2},
  pages={1004--1015},
  year={2022},
  publisher={IEEE}
}

@inproceedings{wang2023inversesr,
  title={Inversesr: 3d brain mri super-resolution using a latent diffusion model},
  author={Wang, Jueqi and Levman, Jacob and Pinaya, Walter Hugo Lopez and Tudosiu, Petru-Daniel and Cardoso, M Jorge and Marinescu, Razvan},
  booktitle={International conference on medical image computing and computer-assisted intervention},
  pages={438--447},
  year={2023},
  organization={Springer}
}

@article{yurt2021mustgan,
  title={mustGAN: Multi-stream Generative Adversarial Networks for MR Image Synthesis},
  author={Yurt, Mahmut and Dar, Salman U. H. and Erdem, Aykut and Erdem, Erkut and Oguz, Kader K. and Cukur, Tolga},
  journal={Medical Image Analysis},
  volume={70},
  pages={101944},
  year={2021},
  doi={10.1016/j.media.2020.101944}
}

@article{isensee2021nnu,
  title={nnU-Net: a self-configuring method for deep learning-based biomedical image segmentation},
  author={Isensee, Fabian and Jaeger, Paul F. and Kohl, Simon A. A. and Petersen, Jens and Maier-Hein, Klaus H.},
  journal={Nature Methods},
  volume={18},
  pages={203--211},
  year={2021},
  doi={10.1038/s41592-020-01008-z}
}

@article{li2025cross,
  title={Cross-Scale Texture Supplementation for Reference-based Medical Image Super-Resolution},
  author={Li, Yinghua and Hao, Weiao and Zeng, Hao and Wang, Longguang and Xu, Jian and Routray, Sidheswar and Jhaveri, Rutvij H. and Gadekallu, Thippa Reddy},
  journal={IEEE Journal of Biomedical and Health Informatics},
  year={2025},
  note={Early Access},
  doi={10.1109/JBHI.2025.3572502}
}

@article{yu2019ea,
  title={Ea-GANs: Edge-Aware Generative Adversarial Networks for Cross-Modality MR Image Synthesis},
  author={Yu, Biting and Zhou, Luping and Wang, Lei and Shi, Yinghuan and Fripp, Jurgen and Bourgeat, Pierrick},
  journal={IEEE Transactions on Medical Imaging},
  volume={38},
  number={7},
  pages={1750--1762},
  year={2019},
  doi={10.1109/TMI.2019.2895894}
}

@article{wu2026unpaired,
  title={Unpaired Volumetric Harmonization of Brain MRI with Conditional Latent Diffusion},
  author={Wu, Mengqi and Yu, Minhui and Jing, Shuaiming and Yap, Pew-Thian and Zhang, Zhengwu and Liu, Mingxia},
  journal={Medical Image Analysis},
  volume={107},
  pages={103849},
  year={2026},
  doi={10.1016/j.media.2025.103849}
}

@article{chartsias2017multimodal,
  title={Multimodal MR synthesis via modality-invariant latent representation},
  author={Chartsias, Agisilaos and Joyce, Thomas and Giuffrida, Mario Valerio and Tsaftaris, Sotirios A},
  journal={IEEE transactions on medical imaging},
  volume={37},
  number={3},
  pages={803--814},
  year={2017},
  publisher={IEEE}
}

@article{dar2019image,
  title={Image synthesis in multi-contrast MRI with conditional generative adversarial networks},
  author={Dar, Salman UH and Yurt, Mahmut and Karacan, Levent and Erdem, Aykut and Erdem, Erkut and Cukur, Tolga},
  journal={IEEE transactions on medical imaging},
  volume={38},
  number={10},
  pages={2375--2388},
  year={2019},
  publisher={IEEE}
}

@article{zhao2025mamba,
  title={Mamba-Enhanced Diffusion Model for Perception-Aware Blind Super-Resolution of Magnetic Resonance Imaging},
  author={Zhao, Xiaoqiang and Yang, Xiaodong and Song, Zhaoyang},
  journal={IEEE Journal of Biomedical and Health Informatics},
  year={2025},
  publisher={IEEE}
}

@article{zhou2022blind,
  title={Blind super-resolution of 3D MRI via unsupervised domain transformation},
  author={Zhou, Hexiang and Huang, Yawen and Li, Yuexiang and Zhou, Yi and Zheng, Yefeng},
  journal={IEEE Journal of Biomedical and Health Informatics},
  volume={27},
  number={3},
  pages={1409--1418},
  year={2022},
  publisher={IEEE}
}

@inproceedings{benisty2025simple,
  title={SIMPLE: Simultaneous Multi-Plane Self-Supervised Learning for Isotropic MRI Restoration from Anisotropic Data},
  author={Benisty, Rotem and Shteynman, Yevgenia and Porat, Moshe and Ilivitzki, Anat and Freiman, Moti},
  booktitle={International Conference on Medical Image Computing and Computer-Assisted Intervention},
  pages={551--561},
  year={2025},
  organization={Springer}
}

@article{lin2023high,
  title={High-resolution 3D MRI with deep generative networks via novel slice-profile transformation super-resolution},
  author={Lin, Jiahao and Miao, QI and Surawech, Chuthaporn and Raman, Steven S and Zhao, Kai and Wu, Holden H and Sung, Kyunghyun},
  journal={IEEE Access},
  volume={11},
  pages={95022--95036},
  year={2023},
  publisher={IEEE}
}

@article{rousseau2006registration,
  title={Registration-based approach for reconstruction of high-resolution in utero fetal MR brain images},
  author={Rousseau, Francois and Glenn, Orit A and Iordanova, Bistra and Rodriguez-Carranza, Claudia and Vigneron, Daniel B and Barkovich, James A and Studholme, Colin},
  journal={Academic radiology},
  volume={13},
  number={9},
  pages={1072--1081},
  year={2006},
  publisher={Elsevier}
}

@article{lyu2020multi,
  title={Multi-contrast super-resolution MRI through a progressive network},
  author={Lyu, Qing and Shan, Hongming and Steber, Cole and Helis, Corbin and Whitlow, Chris and Chan, Michael and Wang, Ge},
  journal={IEEE transactions on medical imaging},
  volume={39},
  number={9},
  pages={2738--2749},
  year={2020},
  publisher={IEEE}
}

@article{de20242024,
  title={The 2024 Brain Tumor Segmentation (BraTS) challenge: glioma segmentation on post-treatment MRI},
  author={de Verdier, Maria Correia and Saluja, Rachit and Gagnon, Louis and LaBella, Dominic and Baid, Ujjwall and Tahon, Nourel Hoda and Foltyn-Dumitru, Martha and Zhang, Jikai and Alafif, Maram and Baig, Saif and others},
  journal={arXiv preprint arXiv:2405.18368},
  year={2024}
}

@article{zhao2020smore,
  title={SMORE: a self-supervised anti-aliasing and super-resolution algorithm for MRI using deep learning},
  author={Zhao, Can and Dewey, Blake E and Pham, Dzung L and Calabresi, Peter A and Reich, Daniel S and Prince, Jerry L},
  journal={IEEE transactions on medical imaging},
  volume={40},
  number={3},
  pages={805--817},
  year={2020},
  publisher={IEEE}
}

@article{benisty2024simple,
  title={SIMPLE: Simultaneous Multi-Plane Self-Supervised Learning for Isotropic MRI Restoration from Anisotropic Data},
  author={Benisty, Rotem and Shteynman, Yevgenia and Porat, Moshe and Illivitzki, Anat and Freiman, Moti},
  journal={arXiv preprint arXiv:2408.13065},
  year={2024}
}

@article{liu20213d,
  title={3D isotropic super-resolution prostate MRI using generative adversarial networks and unpaired multiplane slices},
  author={Liu, Yucheng and Liu, Yulin and Vanguri, Rami and Litwiller, Daniel and Liu, Michael and Hsu, Hao-Yun and Ha, Richard and Shaish, Hiram and Jambawalikar, Sachin},
  journal={Journal of Digital Imaging},
  volume={34},
  pages={1199--1208},
  year={2021},
  publisher={Springer}
}

@article{deoni2022simultaneous,
  title={Simultaneous high-resolution T2-weighted imaging and quantitative T 2 mapping at low magnetic field strengths using a multiple TE and multi-orientation acquisition approach},
  author={Deoni, Sean CL and O'Muircheartaigh, Jonathan and Ljungberg, Emil and Huentelman, Mathew and Williams, Steven CR},
  journal={Magnetic Resonance in Medicine},
  volume={88},
  number={3},
  pages={1273--1281},
  year={2022},
  publisher={Wiley Online Library}
}

@article{uus2025scanner,
  title={Scanner-based real-time three-dimensional brain+ body slice-to-volume reconstruction for T2-weighted 0.55-T low-field fetal magnetic resonance imaging},
  author={Uus, Alena and Neves Silva, Sara and Aviles Verdera, Jordina and Payette, Kelly and Hall, Megan and Colford, Kathleen and Luis, Aysha and Sousa, Helena and Ning, Zihan and Roberts, Thomas and others},
  journal={Pediatric Radiology},
  pages={1--14},
  year={2025},
  publisher={Springer}
}

@article{sui2021gradient,
  title={Gradient-guided isotropic MRI reconstruction from anisotropic acquisitions},
  author={Sui, Yao and Afacan, Onur and Jaimes, Camilo and Gholipour, Ali and Warfield, Simon K},
  journal={IEEE transactions on computational imaging},
  volume={7},
  pages={1240--1253},
  year={2021},
  publisher={IEEE}
}

@article{luo2024target,
  title={Target-guided diffusion models for unpaired cross-modality medical image translation},
  author={Luo, Yimin and Yang, Qinyu and Liu, Ziyi and Shi, Zenglin and Huang, Weimin and Zheng, Guoyan and Cheng, Jun},
  journal={IEEE Journal of Biomedical and Health Informatics},
  year={2024},
  publisher={IEEE}
}

@inproceedings{kim2024adaptive,
  title={Adaptive latent diffusion model for 3d medical image to image translation: Multi-modal magnetic resonance imaging study},
  author={Kim, Jonghun and Park, Hyunjin},
  booktitle={Proceedings of the IEEE/CVF Winter Conference on Applications of Computer Vision},
  pages={7604--7613},
  year={2024}
}

@inproceedings{zhu2017unpaired,
  title={Unpaired image-to-image translation using cycle-consistent adversarial networks},
  author={Zhu, Jun-Yan and Park, Taesung and Isola, Phillip and Efros, Alexei A},
  booktitle={Proceedings of the IEEE international conference on computer vision},
  pages={2223--2232},
  year={2017}
}

@article{jiang2007svr,
  title={MRI of moving subjects using multislice snapshot images with volume reconstruction (SVR): application to fetal, neonatal, and adult brain studies},
  author={Jiang, Shuiping and Xue, Hui and Glover, Andrew and Rutherford, Mary and Rueckert, Daniel and Hajnal, Joseph V.},
  journal={IEEE Transactions on Medical Imaging},
  year={2007},
  volume={26},
  pages={967--980},
  doi={10.1109/TMI.2007.895456}
}

@article{gholipour2010robust,
  title={Robust super-resolution volume reconstruction from slice acquisitions: application to fetal brain MRI},
  author={Gholipour, Ali and Estroff, Judy A. and Warfield, Simon K.},
  journal={IEEE Transactions on Medical Imaging},
  year={2010},
  volume={29},
  pages={1739--1758},
  doi={10.1109/TMI.2010.2051680}
}

@inproceedings{chen2018dcsrn,
  title={Efficient and Accurate MRI Super-Resolution Using a Generative Adversarial Network and 3D Multi-level Densely Connected Network},
  author={Chen, Yao and Shi, Feng and Christodoulou, Aggelos G. and Xie, Yao and Zhou, Zhiming and Li, Debiao},
  booktitle={Medical Image Computing and Computer Assisted Intervention -- MICCAI 2018},
  series={Lecture Notes in Computer Science},
  volume={11070},
  pages={91--99},
  year={2018},
  publisher={Springer},
  doi={10.1007/978-3-030-00928-1_11}
}

@article{pham2019multiscale,
  title={Multiscale brain MRI super-resolution using deep 3D convolutional networks},
  author={Pham, Cong and Ducournau, Alexandre and Fablet, Ronan and Rousseau, Fran{\c{c}}ois},
  journal={Computerized Medical Imaging and Graphics},
  year={2019},
  volume={77},
  pages={101647},
  doi={10.1016/j.compmedimag.2019.101647}
}

@inproceedings{wang2020enhancedgan,
  title={An Enhanced Generative Adversarial Network for 3D Brain MRI Super-Resolution},
  author={Wang, Jiancong and Weyn, Benjamin and Buehler, Patrick and others},
  booktitle={Proceedings of the IEEE/CVF Winter Conference on Applications of Computer Vision (WACV)},
  year={2020},
  pages={712--721},
  doi={10.1109/WACV45572.2020.9093355}
}

@article{lin2023sptsr,
  title={High-Resolution 3D MRI With Deep Generative Networks via Novel Slice-Profile Transformation Super-Resolution},
  author={Lin, Shuo and Lin, Shiqian and Wang, Bin and Liu, Jiacheng and others},
  journal={IEEE Access},
  year={2023},
  volume={11},
  pages={95022--95036},
  doi={10.1109/ACCESS.2023.3307577}
}

@article{remedios2025eclare,
  title={ECLARE: Efficient cross-planar learning for anisotropic resolution enhancement},
  author={Remedios, Samuel W. and Ranjitkar, Pratik and Zaiss, Moritz and Deshmane, Anagha and Chavhan, Govind and Rueckert, Daniel and {\"U}nser, Michael and others},
  journal={arXiv preprint arXiv:2503.11787},
  year={2025}
}

@article{xie2022parallelcyclegan,
  title={Synthesizing high-resolution magnetic resonance imaging using parallel cycle-consistent generative adversarial networks for fast magnetic resonance imaging},
  author={Xie, H. and others},
  journal={Medical Physics},
  year={2022},
  doi={10.1002/mp.15510}
}

@article{pinaya2022ldmbrain,
  title={Brain Imaging Generation with Latent Diffusion Models},
  author={Pinaya, Walter Hugo Lopez and Graham, Mark S. and Gray, Ronald and others},
  journal={arXiv preprint arXiv:2209.07162},
  year={2022}
}

@article{iglesias2023synthsr,
  title={SynthSR: resolution-agnostic MRI synthesis for clinical neuroimaging},
  author={Iglesias, Juan Eugenio and Billot, Benjamin and Greve, Douglas N. and Van Leemput, Koen and Fischl, Bruce and Dalca, Adrian V.},
  journal={Science Advances},
  year={2023},
  doi={10.1126/sciadv.add3607}
}

@article{cohen2018distribution,
  title={Distribution Matching Losses Can Hallucinate Features in Medical Image Translation},
  author={Cohen, Joseph Paul and Luck, Luke and Honari, Sina},
  journal={arXiv preprint arXiv:1808.07299},
  year={2018}
}

@article{wang2004ssim,
  title={Image quality assessment: from error visibility to structural similarity},
  author={Wang, Zhou and Bovik, Alan C and Sheikh, Hamid R and Simoncelli, Eero P},
  journal={IEEE Transactions on Image Processing},
  volume={13},
  number={4},
  pages={600--612},
  year={2004},
  publisher={IEEE}
}

@article{canny1986,
  title={A computational approach to edge detection},
  author={Canny, John},
  journal={IEEE Transactions on Pattern Analysis and Machine Intelligence},
  volume={PAMI-8},
  number={6},
  pages={679--698},
  year={1986},
  publisher={IEEE}
}

\end{document}